%% file: 0_main.tex
\documentclass{article}

\usepackage{microtype}
\usepackage{graphicx}
\usepackage{subfigure}
\usepackage{booktabs} %

\usepackage{hyperref}

\usepackage[accepted]{icml2025}

\usepackage{amsmath}
\usepackage{amssymb}
\usepackage{mathtools}
\usepackage{amsthm}
\usepackage{dashbox}
\newcommand\dboxed[1]{\dbox{\ensuremath{#1}}}
\usepackage{url}
\usepackage[english]{babel}
\usepackage{ulem}
\usepackage{wrapfig}
\usepackage{booktabs}       %
\usepackage{nicefrac}       %
\usepackage{microtype}      %
\usepackage{soul}
\usepackage{multirow}
\usepackage{colortbl}
\sethlcolor{LightBlue2}
\usepackage{mathrsfs}
\usepackage{amsmath,amsfonts,amssymb,amsthm}
\usepackage{resizegather}

\usepackage{placeins}
\usepackage{subcaption}
\usepackage{adjustbox}

\usepackage[capitalize,noabbrev]{cleveref}

\theoremstyle{plain}
\newtheorem{theorem}{Theorem}[section]

\theoremstyle{definition}
\newtheorem{definition}[theorem]{Definition}

\theoremstyle{remark}

\input{paper_macros}
\usepackage[disable,textsize=tiny]{todonotes}

\icmltitlerunning{EigenLoRAx}

\begin{document}

\twocolumn[
\icmltitle{EigenLoRAx: Recycling Adapters to Find Principal Subspaces for Resource-Efficient Adaptation and Inference}

\icmlsetsymbol{equal}{*}

\begin{icmlauthorlist}
\icmlauthor{Prakhar Kaushik*}{yyy}
\icmlauthor{Ankit Vaidya*}{yyy}
\icmlauthor{Shravan Chaudhari}{yyy}
\icmlauthor{Alan Yuille}{yyy}
\end{icmlauthorlist}

\icmlaffiliation{yyy}{Department of Computer Science, Johns Hopkins University, Baltimore, USA}

\icmlcorrespondingauthor{Prakhar Kaushik}{pkaushi1@jh.edu}

\icmlkeywords{Machine Learning, ICML}

\vskip 0.3in
]

\printAffiliationsAndNotice{\icmlEqualContribution} %

\begin{abstract}
\input{0_abstract}
\end{abstract}

\section{Introduction}
Recent advancements in machine learning have driven the rise of large-scale models with billions of parameters. However, the size and complexity of these models not only make it impractical for most researchers to train or fine-tune them on downstream tasks but also contribute significantly to their carbon footprint, raising concerns about environmental sustainability.
To address these challenges, there has been growing interest in parameter-efficient finetuning (PEFT) methods, such as adapters~\citep{pmlr-v97-houlsby19a, Chen2022AdaptFormerAV, luo2023towards}, low rank adaptation (LoRA) methods~\citep{hu2021lora,kopiczko_vera_2023,dora}, prompt-based methods~\citep{tune1, tune2, tune3}.
LoRA and its follow-up works~\citep{meng_pissa_2024,dora} have gained significant attention for their simplicity. This has fueled the proliferation of thousands of low-rank adapters within the growing open-source community.
Given that these adapters are underutilized, an important question arises: Can we recycle the information contained in them to improve the efficiency of subsequent tasks?
Recent work has shown that weight updates in deep neural networks occur within low-dimensional invariant subspaces~\citep{kwon2024efficientcompressionoverparameterizeddeep}, aligning with the universality hypothesis that neural network behavior and learned representations often reside in shared, structured subspaces~\cite{chughtai2023toymodeluniversalityreverse, guth2024on}. This suggests that LoRA adapters may similarly share a \textit{principal subspace} that can be reused, eliminating the need to rediscover it during the training of new adapters.

We introduce \textbf{EigenLoRAx}, a parameter-efficient fine-tuning (PEFT) method that leverages this insight by decomposing the weights of a set of trained adapters into principal components, identifying a compact, information-dense subspace. EigenLoRAx reduces the number of learnable parameters by up to $\mathbf{100\times}$ compared to LoRA, accelerates optimization by up to $\mathbf{2\times}$ for new adapters, and enables more memory-efficient inference with multiple task adapters, particularly benefiting edge devices~\citep{edge}. Additionally, in low-resource domains, we demonstrate that EigenLoRAx can be further enhanced by augmenting the principal subspace with random components, orthogonalized with respect to the existing subspace, preserving its efficiency while retaining performance.

Furthermore, we provide an initial theoretical analysis of EigenLoRAx.
Our experiments across a wide range of vision and language tasks demonstrate its versatility and effectiveness, reinforcing the potential of shared subspaces in neural network adaptation.

\begin{figure*}[!hbt]
\begin{center}
\includegraphics[width=\textwidth]{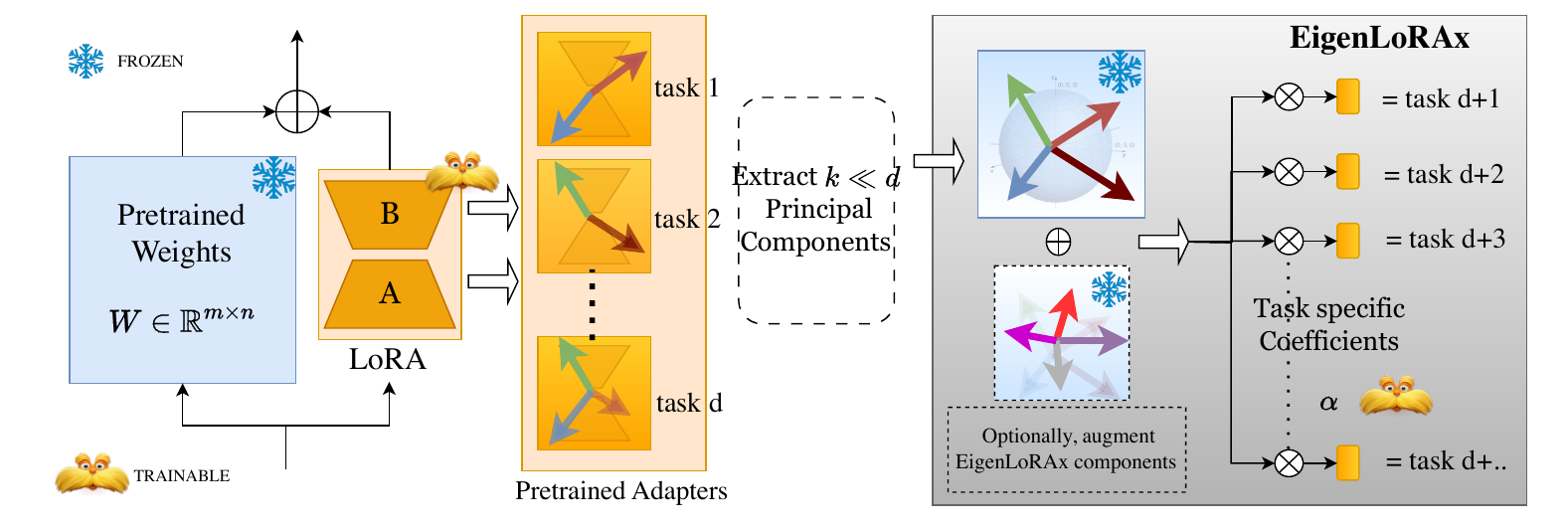}
\caption{\small{LoRA uses low-rank matrices for task-specific finetuning. We observe that LoRA adapters share a principal subspace across task domains. By recycling pretrained adapters, we extract \textit{task-invariant} principal components, enabling efficient representation of both existing and future LoRAs using compact \textit{task-specific} coefficients. This improves training speed, parameter efficiency, and memory usage. In low-resource settings, where pretrained adapters are scarce, we augment the subspace with randomly initialized components, ensuring orthogonality via the Gram-Schmidt process, ensuring they complement the extracted subspace without redundancy.}}
\label{fig:fig1}
\end{center}
\end{figure*}

\autoref{fig:fig1} provides an overview of our method. We introduce \textbf{EigenLoRAx} (ELoRAx), which recycles pretrained adapters by identifying a shared \textit{task-invariant} weight subspace. We hypothesize (and validate experimentally) that task-specific weights lie within this subspace, allowing for more efficient training with fewer parameters. This reduces memory footprint and enhances inference efficiency by enabling simultaneous serving of multiple adapters. EigenLoRAx is among the first to recycle pretrained adapters, replacing many while improving further training efficiency. Our key contributions are as follows:

\begin{itemize}
    \item \textbf{(Training)}: EigenLoRAx uses up to $\mathbf{100\times}$ \textbf{fewer parameters than LoRA} and converges up to $\mathbf{2\times}$ \textbf{faster} than comparable methods, achieving similar or better performance.
    \item \textbf{(Inference)}: EigenLoRAx enhances \textbf{memory efficiency during inference} by approximately $\mathbf{18\times}$ on multiple tasks, reducing the number of switchable parameters between tasks.
    \item \textbf{(Applicability)}: We empirically demonstrate the effectiveness of EigenLoRAx across a wide range 
    , including text and image data, validating the existence of shared principal subspaces across modalities. It also retains performance in \textbf{zero-shot} and \textbf{low resource} scenarios.
    \item \textbf{(Scaling)}: EigenLoRAx can be scaled up to recycle hundreds of underutilized pretrained adapters.
\end{itemize}
\section{Related Works}
\input{2_relatedwork}

\section{Method}
\input{4_definitions}

\input{4_method}

\section{Experiments and Results}\label{sec:experiments}
In this section, we demonstrate the efficacy and versatility of EigenLoRAx across diverse tasks, modalities, and model architectures, highlighting its individual advantages. EigenLoRAx requires significantly fewer parameters to match or surpass LoRA’s performance (Tables~\ref{tab:vision_models}, \ref{tab:glue_benchmark_results}) and achieves similar or faster loss convergence (Figure~\ref{fig:traininglosscola}), making it a cost-effective alternative to random initialization and other methods~\citep{meng_pissa_2024}. Additionally, we showcase its memory-efficient inference capabilities with a Stable Diffusion text-to-image generation model~\citep{diffusion} (Section~\ref{sec:diffusion}). Notably, EigenLoRAx retains its efficiency even in low-resource scenarios where a large number of LoRAs are unavailable.

\paragraph{Note on Baselines} Our focus is on recycling adapter knowledge and improving training and memory efficiency while maintaining performance, not solely on maximizing performance. We compare primarily with LoRA, as EigenLoRAx builds its principal subspace using pretrained LoRA adapters. Using better adapters and optimization could further enhance the subspace and performance. 

Due to lack of space, more experiments (3D Object pose estimation) and detailed ablation experiments are presented in \cref{sec:appendix}. 

\input{3_experiments}

\section{Conclusion}

We introduce EigenLoRAx, a significantly efficient model finetuning and inference method that recycles publicly available pretrained adapters by finding a shared principal subspace. This allows finetuning on new data by simply learning the lightweight coefficients of the shared subspace, and also requires less number of parameters to be saved for new tasks. Our comprehensive and diverse experiments show that EigenLoRAx is applicable to a large range of problems and model architectures. We believe that EigenLoRAx has the potential to mitigate the perpetually widening compute resource gap~\citep{Ahmed2020TheDO, Besiroglu2024TheCD} and reduce the environmental cost of training and using machine learning models~\citep{Wu2021SustainableAE, Ligozat2021UnravelingTH}. It also holds promise for training personalized models~\citep{tan2024democratizing} on low-resource devices, in privacy-critical use-cases. We have a large number of experiments ($6+$) on a diverse set of complex models, tasks and modalities. We have shown that EigenLoRAx excels in faster and efficient learning, memory savings and zero shot performance which differentiates it from conventional PeFT models. While our work is application-focused, we believe it is the first work to hypothesize and empirically prove the existence of shared weight subspaces of neural networks. This important insight has significant implications of model merging, efficiency, mechanistic interpretability, and neural learning theory.

\nocite{sun2025transformersquaredselfadaptivellms, Gain_2020_WACV, Kaushik_2024_CVPR}
\bibliography{iclr2025_conference}
\bibliographystyle{icml2025}

\newpage
\appendix
\onecolumn
\input{8_appendix}

\end{document}

%% file: paper_macros.tex
\newcommand{\tnew}{t_{d+1}}
\newcommand{\Wtrue}{{W^*}_{d+1}}
\newcommand{\Wnew}{W_{d+1}}
\newcommand{\Wegn}{W_{\mathcal{E}}}
\newcommand{\singularsum}{\lVert C\lVert^2_2\sum_{i=K+1}^{nd}\sigma^2_i}
\newcommand{\alphatrue}{\alpha^{*}_{d+1}}
\newcommand{\Vk}{\mathcal{V}^T_{K}}
\newcommand{\V}{\mathcal{V}_{K}}
\newcommand{\alphalearnt}{\alpha_{d+1}}
\newcommand{\Frisk}{R_{\mathcal{S}_{d+1}}^F}
\newcommand{\emprisk}{\hat{R}_{\mathcal{S}_{d+1}}^F}
\newcommand{\Rademacher}{\mathcal{R}_{s_{d+1}}}
\newcommand{\loss}{\mathcal{\ell}_{\mathcal{S}_t}^F}
\newcommand{\hegn}{h^{\mathcal{E}}}
\newcommand{\htrue}{h^{*}}

\newcommand{\Xnew}{X_{d+1}}
\newcommand{\Ynew}{Y_{d+1}}

%% file: 0_abstract.tex
The rapid growth of large models has raised concerns about their environmental impact and equity in accessibility due to significant computational costs. Low-Rank Adapters (LoRA) offer a lightweight solution for finetuning large models, resulting in an abundance of publicly available adapters tailored to diverse domains. We ask: Can these pretrained adapters be leveraged to further streamline adaptation to new tasks while addressing these challenges? We introduce EigenLoRAx, a parameter-efficient finetuning method that recycles existing adapters to create a principal subspace aligned with their shared domain knowledge which can be further augmented with orthogonal basis vectors in low-resource scenarios. This enables rapid adaptation to new tasks by learning only lightweight coefficients on the principal components of the subspace - eliminating the need to finetune entire adapters. EigenLoRAx requires significantly fewer parameters and memory, improving efficiency for both training and inference. Our method demonstrates strong performance across diverse domains and tasks, offering a scalable for edge-based applications, personalization, and equitable deployment of large models in resource-constrained environments, and is arguably the most efficient finetuning method, till date. The code is available \href{https://github.com/toshi2k2/EigenLoRA/}{here}.

%% file: 2_relatedwork.tex
Low-Rank Adaptation refers to modeling neural network weight updates as a function of low-rank matrices instead of training the entire weight matrix.
This is a well-established line of research starting from Burer-Monteiro factorization~\citep{Burer2003ANP}, with a recent resurgence by~\citet{hu2021lora} (LoRA), who used it as a technique to finetune LLMs; and other related variants ~\citep{Ma2024, chi19_low, kwon2024efficientcompressionoverparameterizeddeep}. 
However, with rapid growth in the scale of models, Low-Rank Adaptation has also become relatively expensive; for example, LoRA with a rank of 16 on GPT-3~\cite{browngpt3} requires 75.5 million parameters. Consequently, more efficient low-rank fine-tuning methods are being developed. Mixture of experts models~\citep{huang2023lorahub, wu2024mixture, diao_mixture--domain-adapters_2023, zhong2024multi, zhou2018x} have been proposed as a method to adapt to new domains using a mixture of low-rank modules. But these approaches typically require a substantial number of high-quality adapters to work efficiently~\citep{acl-2024-long}, which can significantly increase the model memory requirements~\citep{zhou2022mixtureofexperts}. Furthermore, complex gating or weighting mechanisms utilized with these models can exhibit training instability~\citep{zoph2022stmoedesigningstabletransferable}. 

Recent methods have aimed to learn better subspaces for low-rank optimization, primarily by decomposing model weights into singular vectors for improved training. \citet{meng_pissa_2024} demonstrate that initializing LoRA with singular vectors is superior to random initialization, while ~\citet{sharma_laser_2023} find that removing minor singular components enhances robustness. Using randomly initialized principal components~\citep{kopiczko_vera_2023} or weight matrices~\citep{nola} has also been explored to reduce the number of trainable parameters. However, as shown in Section~\ref{sec:experiments}, random initialized subspaces may not be very useful. This is intuitive as the random subspace may not have an overlap with domain-specific principal subspaces. On the other hand, EigenLoRAx uses trained adapters to extract a \textit{principal subspace} suitable for a given domain of tasks resulting in a better subspace initialization than and parameter efficiency. 
Given our focus on resource and computation efficiency in this work, we focus primarily on LoRA~\citep{hu2021lora} as our main baseline, but EigenLoRAx can be used with any PEFT method like~\cite{dora, zhang2023adaloraadaptivebudgetallocation} where task-specific weights can be analyzed together. 

This work is a direct application of \textbf{the Universal Weight Subspace Hypothesis}~\cite{kaushik2025universalweightsubspacehypothesis}, for creating an efficient PEFT method focused on low rank adaptation.

%% file: 4_definitions.tex
In this section, we present the theoretical foundation~\cref{sec:theory} and algorithmic details~\cref{sec:algo} of our method, followed by a discussion on hyperparameter selection and an assessment of its practical advantages. Note that we use the terms EigenLoRA, EigenLoRAx, ELoRA and ELoRAx interchangeably.

\subsection{Theoretical Preliminaries}
\label{sec:theory}

For a full rank weight matrix $W \in \mathbb{R}^{m \times n} $ that learns to map input space $X\in\mathbb{R}^{m}$ to output space $\mathbb{R}^{n}$, the rank is expressed as $\min(m,n)$. As the rank of $ W $ increases, modifying it to accommodate new tasks becomes computationally expensive and increasingly complex. This is a common challenge faced when finetuning pretrained large foundation models. LoRA is a parameter efficient finetuning approach used for large pretrained models with weights $W_0$ that mitigates this challenge by merely learning low-rank weight updates $W$ such that the risk between $Y$ and $W_0X + WX + b$ is minimized. Instead of directly learning $W$, LoRA proposes to learn a lower ranked decomposition of $W$ by learning two low-rank matrices, $ B \in \mathbb{R}^{m \times r} $ and $ A \in \mathbb{R}^{r \times n} $, both having ranks $r$.  This factorization ensures that the product $ BA $ retains the original dimensions of $ W_0 $ while having a significantly reduced rank.
As a result, although the transformation defined by $ BA $ maps from $ \mathbb{R}^m $ to $ \mathbb{R}^n $, it does not span the full space of such mappings due to its constrained rank. The low-rank weight matrices result in substantially smaller number of trainable parameters than the full rank parameter count of $m\cdot n$. Such parameter efficient finetuning makes LoRA a computationally viable alternative for fine-tuning large-scale models.    

Previous works such as \cite{meng_pissa_2024, dora} have proposed the existence of a common parameter subspace implying the idea of shared principal subspace. We highlight that LoRA adapters share such a lower dimensional shared principal subspace when finetuned for diverse tasks. Along with reduction in computational overhead, it reinforces the idea that task-relevant transformations reside within a compact, reusable subspace. To formalize this, we first define a space of tasks representable by linear transformation matrices, providing a foundation for analyzing the role of shared principal subspaces in model adaptation.

\begin{definition}[Task definition for LoRAs] 
\label{def:lineartasks}
We first define a LoRA task $t_i (X_i,Y_i): \mathbb{R}^m \rightarrow \mathbb{R}^n$ such that $Y_i = W^*_iX_i + b$ where $b$ is some constant. Then the LoRA task domain $\mathcal{T}_d$ is a set of $d$ such tasks, $\mathcal{T}_d = \{t_i\}^{d}_{i=1}$.
\end{definition}

For a given set of pretrained weights (such as those from a foundation model) $W_0\in\mathbb{R}^{m\times n}$, LoRA weights $BA$ at any layer modify the output as $W_0X + BAX + \epsilon_t$, allowing the model to adapt to the new task and converge toward the optimal solution $W^*_t$. The key here is that only $B$ and $A$ weights are updated during finetuning. 
Without loss of generality, 
assume $r\ll n$ and let the true transformation matrix $W^*_t \in \mathbb{R}^{r \times n}$ be interpreted as $r$ $n$-dimensional vectors: $\mathbf{w}^{*1}_t, ..., \mathbf{w}^{*r}_t \in \mathbb{R}^{n}$. Finding LoRA weights is equivalent to finding sets of these $r$ vectors in $\mathbb{R}^{n}$. 

\begin{definition}[Set of LoRA weights]
\label{def:loraset}
We define the weights of a LoRA adapted for task $t_i$ as $B_iA_i$. Both $B_i$ and $A_i$ will have their own individual subspaces. For the purpose of the analysis we will consider a generic task specific weight matrix $W_i \in \mathbb{R}^{m\times n}$ adapted to task $t_i$ such that $n<m$ and its rank $r<n$. The analysis, however, is valid for both $B_i$ and $A_i$. 
Now can define a set of LoRAs as stacked (along columns) weight matrices $\hat{W} = \{W_i \}^{d}_{i=1}$ where each $W_i$ is adapted for a task $t_i\in\mathcal{T}_d$ and a training set $\mathcal{S}_i=\{\{x,y\} \mid x\in X_t, y\in Y_t\}$ where the size of the training is $s_i = |\mathcal{S}_i|$. For theoretical analysis we assume that each training set $X_i \times Y_i$ is distributed according to some unknown Gaussian distribution with mean $\Bar{X_i}$ and $\lVert X_i\lVert_F\leq M$ for some constant $M>0$. Each weight matrix can have different ranks and the following method and analysis will still hold, however, for brevity we assume all weight matrices stacked in $\hat{W}$ to have the same rank $r$. 
\end{definition}
\begin{definition}[Subspace spanned by LoRAs from a task domain $\mathcal{T}_{d}$]
\label{def:subspace}
We define the subspace of weights $\mathcal{Z}_{d} = \{C\hat{W} \mid C\in\mathbb{R}^{m\times m}\}$ spanned within $\mathbb{R}^{m\times n}$. 
\end{definition}
Using Singular Value Decomposition (SVD) or Principal Component Analysis (PCA for a zero-centered $\hat{W}$) , we can obtain $\hat{W}=\mathcal{U}\Sigma \mathcal{V}^T$. We then represent top $K$ right singular vectors of $\hat{W}$ (or top $K$ principal components if $\hat{W}$ is zero-centered) as $\Vk \in\mathbb{R}^{K\times n}=\{\Vk\in\mathbb{R}^{1\times n}\}^K_{k=1}$.
\begin{definition}[Shared principal subspace of LoRAs finetuned in domain $\mathcal{T}_d$]
\label{def:principalsubspace}
We define the shared principal subspace of weights for a task domain $\mathcal{T}_d$ as $\mathcal{Z}^{K}_d = \{\alpha \Vk \mid \alpha\in\mathbb{R}^{m\times K}\}$ spanned by top K principal components of the LoRAs within $\mathbb{R}^{m\times n}$. 
\end{definition}

Next, we introduce the idea of defining a new related task $\tnew$
\begin{definition}[New related task $\tnew$].
\label{def:newtask}
    A new linear task $\tnew$ with true solution $\Wtrue$ is said to be related if it is spanned by the basis of $\hat{W}$ i.e. $\Wtrue = C\hat{W}$ and it holds that $\lVert \Wtrue - \alphatrue\Vk\lVert^2_F \leq \Vert\Wtrue - \alphalearnt\Vk\lVert^2_F$ for all rank $K$ linear transformation matrices $\alphalearnt$ and $\lVert \Wtrue - \alphatrue\Vk\lVert^2_F \leq \singularsum$ where $\sigma_i$'s are singular values of $\hat{W}$. For such a task, we learn coefficients of $K$ principal components $\alphalearnt \in \mathbb{R}^{m\times K}$ resulting in EigenLoRAx weights $\Wegn = \alphalearnt\Vk$. \\
\end{definition}

Definition \ref{def:newtask} establishes a bound over the related-ness of a new task with those in the known task domain $\mathcal{T}_d$. If the true solution of the new task lies majorly in the principal subspace of $\mathcal{T}_d$ i.e. has major principal components (PCs) within the top $K$ principal components of $\hat{W}$ with some finite bound on the misalignment along the PCs orthogonal to the top $K$ PCs of $\hat{W}$, then we can ideally quantify the relation between a new task and a task domain. Any task that has its true solution within a subspace defined by the PCs orthogonal to the top $K$ PCs of $\hat{W}$ is not as closely related as a task with its solution completely or majorly within the principal subspace. A task that has its solution completely orthogonal to all the PCs of $\hat{W}$ is completely unrelated and is not the main focus of this study. 

Next, we present an algorithm to find the principal subspace of the trained adapters and our experiments in Section~\ref{sec:experiments}.

\begin{algorithm}[!htb]
\caption{\textbf{EigenLoRAx PCs} Calculation}
\label{algo:eigenlora}
\begin{algorithmic}
    \STATE {\bfseries Input:} 
    LoRA matrices 
    $\{W_t \in \mathbb{R}^{m \times n}\}_{t=1}^d$ 
    
     , number of PC ($K$), number of pseudo-PC ($P$)\\
    
    \STATE {\bfseries Output:} EigenLoRAx PCs $\Vk$

    \STATE $\hat{W} = \begin{bmatrix*} W_{1}\in \mathbb{R}^{m\times n} & \text{...} & W_{d}\in \mathbb{R}^{m\times n}\end{bmatrix*}$, \COMMENT{Stack LoRA matrices}\\
    
    \STATE Compute the mean of each feature: $ \Bar{W} = \frac{1}{n} \sum_{i=1}^{n} W_i $
    \STATE Subtract the mean: $ \hat{W}_c = \hat{W} - \Bar{W} $

    \STATE Perform SVD: $ \hat{W}_c = U \Sigma V^T $
    
    \STATE Extract the top $ K $ principal components
    \STATE Select the first $ K $ columns of $ \mathcal{V} $: $ \V = V[:, 1:K] $
    
    \STATE Optionally, augment the subspace with $P$ pseudo-PCs
    \FOR{$ p = 1 $ to $ P $}
        \STATE Sample a random vector $ v_p \sim \mathcal{N}(0, I_n) $  \COMMENT{Sample from a normal distribution}
        \STATE Orthogonalize $ v_p $ against all PCs in $ \V $ using Gram-Schmidt:
        \FOR{$ i = 1 $ to $ K+p-1 $}
            \STATE $ v_p = v_p - \frac{v_p^T V_K[:, i]}{\|\V[:, i]\|^2} \V[:, i] $
        \ENDFOR
        \STATE Normalize $ v_p $: $ v_p = \frac{v_p}{\|v_p\|} $
        \STATE Append $ v_p $ to $ \V $ if $ v_p $ is not a null vector
        \STATE $K = K + 1$
    \ENDFOR

    \STATE return $\V, \mu$
    
\end{algorithmic}
\end{algorithm}

\subsection{Algorithm}
\label{sec:algo}

Assume we have $N$ LoRA adapters, each consisting of a set of $A, B$ matrix pairs for every layer, trained on various tasks within a domain $\mathcal{T}_d$ for a given base pretrained model. Algorithm~\ref{algo:eigenlora} computes a list of top $K$ principal components—referred to as EigenLoRAx PCs—that define an initial principal subspace for this domain. 

To construct this subspace, the algorithm aggregates LoRA matrices across tasks for each layer, separately for $A$ and $B$ matrices (though it can also be applied to the product $BA$). Each LoRA matrix, having rank r, is treated as a list of vectors, and a decomposition is performed on this stacked set of vectors. The most significant components extracted from this process serve as a basis for the principal subspace, providing an efficient representation that can be linearly combined to approximate the original LoRA weight matrices. We showcase our algorithm using representative weight matrices $W_t$, where each $W_t$ represents a single $A$ or $B$ matrix from a single LoRA layer of the neural network. In practice, this procedure is applied to all relevant layers.

Since real-world scenarios often involve low-resource domains with limited availability of LoRA adapters, we extend our subspace by introducing additional pseudo-PCs. Specifically, we sample random vectors of the same dimension as each PC and orthogonalize them with respect to all existing PCs. This process can be iterated to generate more pseudo-PCs, thereby augmenting the principal subspace. As empirically shown in Table~\ref{tab:glue_low}, this augmentation strategy significantly outperforms naive random selection of PCs for subspace expansion.

\paragraph{Learning new tasks}\label{method:learningnewelora} Having extracted a set of PCs (including pseudo-PCs, if needed), $\mathcal{V}_K \in\mathbb{R}^{K\times n}=\{\V\in\mathbb{R}^{1\times n}\}^K_{k=1}$, we can approximate a given (LoRA) weight matrix by minimizing $\lVert W - \alpha\Vk\lVert_F$ where $\alpha$ are linear coefficients~\cref{sec:theory}. In fact, we can analytically compute of the given LoRA matrices by calculating the linear coefficients which minimizes the above objective. For new tasks however, for which we do not have a LoRA matrix, we freeze the EigenLoRAx PCs and randomly initialize the $\alpha$s. The forward pass in layer is calculated as 
\begin{align}
    h = W_0x + \dboxed{\alpha_B^T \mathcal{V}_B \alpha_A^T \mathcal{V}_A(x)}.
\end{align}
Here, $W_0$ are the pretrained weights of the base model and $\mathcal{V}_B, \mathcal{V}_A$ are EigenLoRAx components (which represent the shared subspace) that are frozen during training. The corresponding lightweight coefficients $\alpha_B$ and $\alpha_A$ are learned. This reduces the number of learnable parameters from $O(2rn)$ to $O(2K)$, by a factor of $\frac{rn}{K}$ (assuming $\alpha$ to be scalar). 

\phantomsection

%% file: 4_method.tex
Using the definitions \ref{def:lineartasks}, \ref{def:loraset}, \ref{def:newtask}, \ref{def:principalsubspace} and \ref{def:subspace} we state the following theorem; 
\begin{theorem}
\label{thm:2}
    For a task $\tnew$, we assume a hypothesis $h\in\mathcal{H}_{\Wnew}$ expressed as $h(\Wnew,X)=\Wnew \Xnew+W_0\Xnew+b$ where $\Wnew$ has rank $m$, $b$ is some constant and $W_0$ represents weights of a pretrained foundation model that is frozen during finetuning respectively. We have $\hegn\in\mathcal{H}_{\Wegn}, \htrue\in\mathcal{H}_{\Wtrue}$ such that $\hegn(\Wegn,\Xnew) = \alphalearnt\Vk \Xnew+W_0\Xnew+b$ where $\Wegn$ has rank $K$ and $\htrue(\Wtrue,\Xnew) = C\hat{W} \Xnew + W_0\Xnew+b$ where $\htrue(\Wtrue,\Xnew)=\Ynew$ is the true solution for task $\tnew$. For a Lipschitz continuous loss ($\loss(h)$) that is strong convex within the shared principal subspace spanned by principal components $\Vk$  with some Lipschitz constant ($L$), the risk can be written as $\Frisk(h_W) = E_{\mathcal{S}_t}[\loss(h) ]$ , and using Rademacher complexity bounds we can say with probability at least $1-4\delta$ for some $\delta>0$,
    \begin{equation}
    \label{equation:eq2}
        \Vert \Wtrue - \Wnew\lVert^2_F \leq C_1\cdot\Bigg(\frac{\sqrt{m}}{\sqrt{s_t}} \Bigg) + C_2
    \end{equation}
    \begin{equation}
    \label{equation:eq1}
        \Vert \alphatrue\Vk - \Wegn\lVert^2_F \leq C_1\cdot\Bigg(\frac{\sqrt{K}}{\sqrt{s_t}} \Bigg) + \singularsum + C_2
    \end{equation}
    where $\sigma_i$ are singular values of $\hat{W}$, $C$ is some constant such that $\Wtrue=C\hat{W}$ and $C_1$, $C_2$ are some constants.

\begin{proof}
    The derivation is straightforward, we can write the difference in risks for $\hegn$ and $\htrue$ as 
    \begin{align*}
        \Frisk(\hegn) - \Frisk(\htrue) = \mathbb{E}_{\mathcal{S}_t}\big[\loss(\hegn) - \loss(h^*)\big]
    \end{align*}
    By definition of strong convex loss function for some constant $\mu\geq0$, 
    \begin{align*}
        \mathbb{E}_{\mathcal{S}_t}\big[\loss(\hegn) - \loss(h^*)\big] \geq \frac{\mu}{2}\lVert \Wegn - \Wtrue\Vert^2_F
    \end{align*}
    We also know from generalization error bounds using Rademacher Complexity from \cite{bartlett2003rademacher} that with probability at least $1-2\delta$, 
    \begin{align*}
        |\Frisk(\hegn) - \emprisk(\hegn)| \leq \frac{\Rademacher(\mathcal{H}_{\Wegn})}{2} + \sqrt{\frac{\ln(1/\delta)}{2s_t}}  
    \end{align*}
    We can rewrite risk as
    \begin{align*}
        \Frisk(\htrue) - \Frisk(\hegn) =& \Frisk(\htrue) - \emprisk(\htrue)\\ 
        &- \Frisk(\hegn) + \emprisk(\hegn)\\
        &+ \emprisk(\htrue) - \emprisk(\hegn)
    \end{align*}
    Since we know by definition of $\htrue$ that $\emprisk(\htrue) \leq \emprisk(\hegn)$, we can say
    \begin{align*}
        \Frisk(\htrue) - \Frisk(\hegn) \leq \Frisk(\htrue) - \emprisk(\htrue)\\ 
        - \Frisk(\hegn) + \emprisk(\hegn)
    \end{align*}
    
    Then we take a union bound to conclude that with probability at least $1-4\delta$,
    \begin{align*}
        \Frisk(\htrue) - \Frisk(\hegn) \leq \frac{\Rademacher(\mathcal{H}_{\Wegn})}{2} + \sqrt{\frac{2\ln(1/\delta)}{s_{d+1}}}\\ 
        + \frac{\Rademacher(\mathcal{H}_{\Wtrue})}{2}
    \end{align*}
    Hence, we can also say that with probability at least $1-4\delta$,
    \begin{equation}
    \label{eq:totalerror1}
        \frac{\mu}{2}\lVert\Wtrue - \Wegn\lVert^2_F
        \leq \frac{\Rademacher(\mathcal{H}_{\Wegn})}{2} + \sqrt{\frac{2\ln(1/\delta)}{s_t}} + \frac{\Rademacher(\mathcal{H}_{\Wtrue})}{2}
    \end{equation}
    The Rademacher complexity of a low-rank weight matrix class $\mathcal{H}_{\Wegn}$ with rank $K$ can be directly bounded using results from \cite{bartlett2003rademacher} as
    \begin{align*}
        \Rademacher(\mathcal{H}_{\Wegn}) &= \mathcal{O}(\frac{\sqrt{K}\lVert \Wegn \lVert_F}{\sqrt{s_t}})\\ 
    \end{align*}
    We can separate the constants including $\Rademacher(\mathcal{H}_{\Wtrue})$ from \ref{eq:totalerror1} and assume that, for a normalised $\lVert\Wegn\lVert$, it is usually bounded, then we can write:
    \begin{equation}
    \label{eq:totalerror2}
         \lVert\Wtrue - \Wegn\lVert^2_F
        \leq C_1\cdot \Bigg(\frac{\sqrt{K}}{\sqrt{s_t}} \Bigg) + C_2
    \end{equation}
    Similarly, we can also say for $\Wnew$ that
    \begin{equation}
    \label{eq:totalerror}
        \lVert\Wtrue - \Wnew\lVert^2_F
        \leq C_1\cdot\Bigg(\frac{\sqrt{m}}{\sqrt{s_t}}\Bigg) + C_2
    \end{equation}
    This proves \ref{equation:eq2}. Now to further prove \ref{equation:eq1}, we use  properties of Frobenius norm,
    \begin{align*}
        \lVert \Wegn - \alphatrue\Vk \lVert^2_F - \lVert \Wtrue - \alphatrue\Vk \lVert^2_F\\ \leq \Vert \Wegn - \Wtrue\lVert^2_F
    \end{align*}
    Then following from the definition of $\Wtrue$, we can say that,
    \begin{align*}
        \lVert \Wegn - \alphatrue\Vk \lVert^2_F - \singularsum \leq \Vert \Wegn - \Wtrue\lVert^2_F
    \end{align*}
    Finally, using the Rademacher complexity bound we provided earlier, 
    we can say that with probability at least $1-4\delta$
    \begin{align*}
        \lVert \alphatrue\Vk &- \Wegn \lVert^2_F \leq  \lVert\Wtrue - \Wegn\lVert^2_F\\
        &\leq C_1\cdot\Bigg(\frac{\sqrt{K}}{\sqrt{s_t}} \Bigg) + \singularsum + C_2
    \end{align*}
    We can just rewrite $\Wegn = \alphatrue\Vk$ and get the same bound as above for $\lVert \alphatrue - \alphalearnt\lVert^2_F$.
    We can similarly obtain the upper bound for \ref{equation:eq1}
    
    This concludes the proof. 
\end{proof}
\end{theorem}

Theorem \ref{thm:2} provides an upper bound on the Frobenius norm of the difference between $\Wnew$ or $\Wegn$ and the optimal solution $\Wtrue$. \ref{equation:eq1} provides a tighter upper bound on the norm of the difference when task $\tnew$ majorly lies in the shared principal subspace. The extent to which task $\tnew$ lies in the shared principle subspace is captured by the second term involving the sum of squared truncated singular values $\hat{W}$. Hence, if the task completely or majorly lie in the shared principal subspace, then the first term (sqrt(rank)) will dominate the upper bound.  Hence, if rank($\Wnew\geq K$), then we can see that the upper bound in eq. \ref{equation:eq1} will be tighter than in eq. \ref{equation:eq2} where the task lies majorly in the shared principal subspace. Similarly, when $m\leq K$, the upper bound on the difference norm will be tighter for $\Wnew$ than $\Wegn$.  When $\Wtrue$ has a significant alignment or projection along the singular vectors orthogonal to the ones with top $K$ singular values, then the second term in \ref{equation:eq2} comes into picture and it becomes difficult to directly compare the bounds in \ref{equation:eq2} and \ref{equation:eq1}. However, if majority of the variance of $\Wtrue$ is along the singular vectors orthogonal to the top $K$ components, it follows that $\Wegn$ will never be able to achieve convergence while $\Wnew$. In contrast, $\Wnew$ could perform significantly better, as it is not restricted to learning only along the top $K$ principal components of $\hat{W}$. While the assumption that $\Wtrue$ is spanned by the principal components of the shared principal subspace might appear to be very strong, we empirically observe in table \ref{tab:lola} that such an assumption is not impractically far from reality. Particularly, we observe in table \ref{tab:glue_benchmark_results} that for GLUE benchmark, LoRA adapters trained on 5 diverse tasks shared a principal subspace. We see that EigenLoRAx was able to leverage the principal components of this shared subspace with just 12K training parameters learned for a new 6th task and achieve competitive performance compared to fine-tuning full rank weights with 125M parameters or individual LoRA adaptors with 1.2M parameters, even outperforming them in certain tasks. Similarly, table \ref{tab:lola} demonstrates zeroshot performance using only top $K$ principal components of the shared subspace obtained through 500 LoRA adaptors trained on diverse tasks. This further suggests that increasing the number of LoRA adapters enables a richer set of top principal components, effectively spanning the shared subspace and providing broader coverage for new tasks.

\begin{figure}
  \begin{center}
    \includegraphics[width=0.35\textwidth]{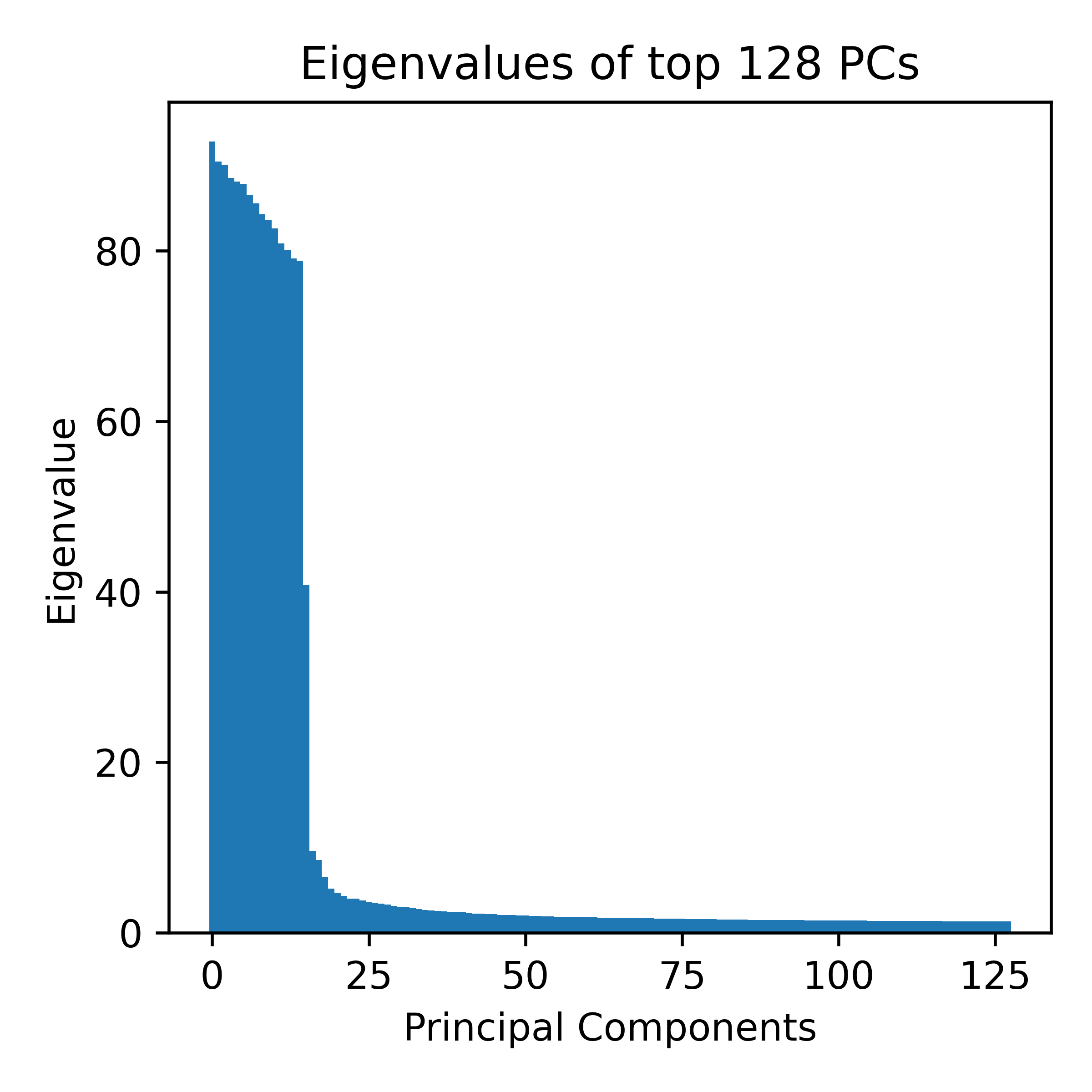}
  \end{center}
  \caption{\small{The top 16 components contain the most information from a total of 4000+ components for $~500$ LoRAs. ($A$ matrices from layer 1 of Mistral-7b model, Lots of LoRAs, see Section~\ref{sec:lotsofloras}).}}
  \label{fig:singular}
\end{figure}
\paragraph{How to choose optimal number of PCs $K$} The hyperparameter $K$, which determines the number of top PC, can be viewed as a function of task domain complexity—simpler domains require a smaller $K$, while more complex domains benefit from a larger $K$.
In practice, we determine $K$ based on empirical observations (\cref{sec:ablation}), evaluating performance across different values. Additionally, we can leverage established techniques from literature, such as explained variance and singular value thresholds \cite{gavish2014optimalhardthresholdsingular}. As illustrated in \cref{fig:singular}, most of the relevant information is often concentrated in a few top EigenLoRAx PCs, providing a practical criterion for selecting $K$.

\paragraph{Memory Efficiency and Complexity} Our method demonstrates significant memory efficiency across experiments. A single set of EigenLoRAx PCs, combined with lightweight task-specific coefficients, can effectively replace both past and future LoRAs within a task domain. This is particularly advantageous when serving a large number of adapters, where frequent loading and unloading in VRAM incurs high latency, or keeping all adapters in memory demands excessive VRAM.
For $d$ LoRAs of rank $r$ and $l$ layers, the memory footprint is $O(2drln)$. For EigenLoRAxs, it is $O(2Kl (d + n) )$. As $r, K \ll n$, EigenLoRAx becomes more memory efficient in terms of memory required to save the models as $d$ increases. This becomes significantly useful for edge devices and large scale user serving AI systems. 

\paragraph{Why does the orthogonal principal component augmentation works?}
Minimum Hyperspherical Energy

%% file: 3_experiments.tex
\begin{table}[!h]
    \caption{Image classification with Vision Transformer. ZS refers to zero-shot. AUG refers to Augmented for Low-Resource. EigenLoRAx matches or increases performance with drastically fewer number of parameters.}
    \begin{center}
    \begin{sc}
    \begin{adjustbox}{width=\columnwidth}
    \begin{tabular}{lcccc}
        \toprule
         & \# Train & CIFAR & Food & Flowers \\
         & Params  & 100 & 101 & 102 \\
        \midrule
        Full Training & 86M  & 97.0 & 96.64 & 98.82 \\
        Base Model &  15K & 90.07 & 90.8 & 80.71 \\
        LoRA ($r=4$)  & +147K & 93.79 & \textbf{95.73} & 95.03  \\
        LoRA ($r=1$)  & +36K  & 92.45 & 91.07 & 90.14  \\
        VeRA & +18K  & 90.87 & 91.75 & 91.25 \\
        \rowcolor{gray!20} ELoRAx$^{\text{AUG}}$ & +1K & 94.4  & 95.01 & 97.5 \\
        \rowcolor{gray!20} ELoRAx & +\textbf{96}  & \textbf{94.8} & 95.14 & \textbf{98.44} \\
        \rowcolor{gray!20} ELoRAx$^{\text{ZS}}$ & +0& 91.4  & 92.48 & 95.7 \\
        \bottomrule
    \end{tabular}
    \label{tab:vision_models}
    \end{adjustbox}
    \end{sc}
    \end{center}
\end{table}
\begin{table*}[!hbt]
    \caption{\small{GLUE benchmark results. We report Matthew's correlation for CoLA, Pearson correlation for STS-B, and accuracy for the remaining tasks. In all cases, higher values indicate better performance.
}}
    \begin{center}
    \begin{sc}
    \begin{tabular}{lcccccccccc}
        \hline
        & \# Trainable & \multirow{2}{*}{MRPC} & \multirow{2}{*}{SST-2} & \multirow{2}{*}{CoLA} & \multirow{2}{*}{QNLI} & \multirow{2}{*}{RTE} & \multirow{2}{*}{STS-B} & \multirow{2}{*}{Avg.} \\
        Method & Parameters & & & & & & & \\
        \hline
        Full Training  & 125M & 88.97 & 91.28 & 59.81 & 92.29 & 79.78 & 90.89 & 83.84 \\
        PISSA [\citenum{meng_pissa_2024}] & 1.2M  & 86.52 & 94.15 & 61.32 & 92.15 & 71.84 & 90.25 & 82.70\\
        \rowcolor{gray!20} EigenLoRAx$^\text{init}$ & 1.2M & 89.71 & 93.35 & 61.58 & 92.2 & 74.73 & 89.56 & 83.52 \\
        \hline
        LoRA ($r=32$) & 1.2M & 86.76 & \textbf{94.72}& 59.56 & 92.53 & 77.61 & \textbf{90.81 }& \textbf{83.67} \\
        VeRA ($r=256$) & 25K  & 75.98 & 93.23 & 54.14 & 89.21 & 66.78 & 87.03 & 77.72 \\
        \rowcolor{gray!20} EigenLoRAx & \textbf{12K} & \textbf{87} & 94.15& \textbf{59.81}& \textbf{92.73}& \textbf{77.62}& 90.58 & 83.65 \\
        \hline
    \end{tabular}
    \label{tab:glue_benchmark_results}
    \end{sc}
    \end{center}
\end{table*}
\subsection{Image Classification}~\label{sec:img_class}
This simpler task involves related datasets where the LoRAs used to construct EigenLoRAx are well-aligned with the downstream tasks, highlighting its finetuning efficiency.
\paragraph{Setup} We evaluate EigenLoRAx using a pretrained Vision Transformer (ViT)~\cite{vision_transformer} across 3 datasets. Each dataset is partitioned into 5–6 non-overlapping sub-datasets, mimicking continual learning~\cite{kaushik2021understandingcatastrophicforgettingremembering} and federated learning~\cite{federated} setups. As the sub-datasets are derived from the same source, their tasks are more domain-aligned.
For EigenLoRAx, we compute principal components (PCs) using all but one LoRA trained on individual sub-datasets (leave-one-out approach, Algorithm~\ref{algo:eigenlora}). The coefficient matrix $\alpha$ for the excluded task is then learned as described in Section~\ref{sec:algo}. All methods are finetuned for 10 epochs, with additional details in Appendix~\ref{appendix:hyperparameters}.
\paragraph{Parameter Efficiency} \autoref{tab:vision_models} summarizes our experimental results. All models require training the last linear layer (approx. 15K parameters) due to the pre-trained ViT having a different number of categories. For the Base Model, no additional parameters are trained. EigenLoRAx adapts to new sub-datasets using only two principal components (96 additional parameters), enabling it to match or outperform LoRA and VeRA, which use significantly more parameters. We also tested a zero-shot EigenLoRAx (weight initialized randomly within the principal subspace), training only the last layer. This model outperforms the base model with no additional parameters, demonstrating the effectiveness of principal subspace extraction. We also test a low resource scenario (ELoRAx$^{\text{AUG}}$), where only 2 LoRAs are available for extracting the PCs, which are then augmented using random, orthogonal PCs as described in \cref{algo:eigenlora}.
\begin{figure*}[!h]
\label{fig:convergence}
  \centering
  \includegraphics[width=\textwidth]{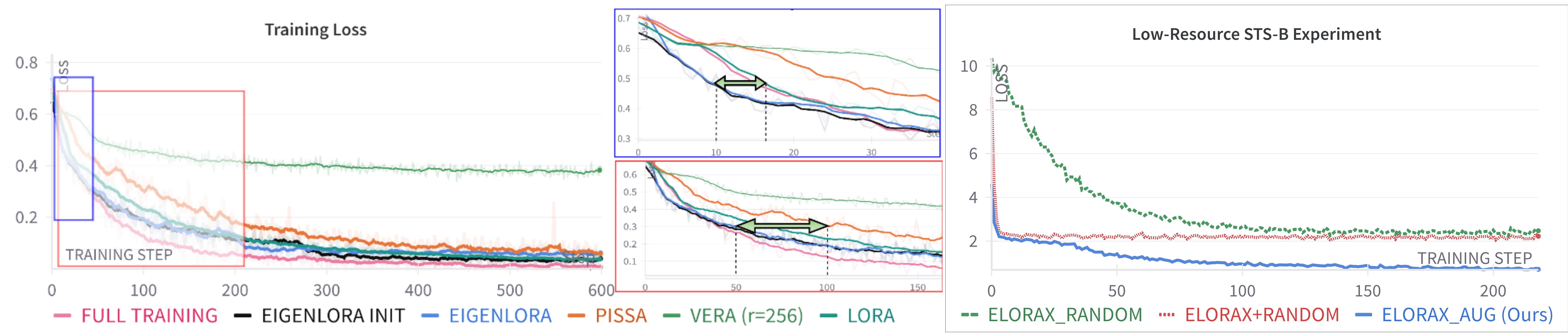}
\caption{
\small{\textbf{Fast Convergence and Better Initialization} (left) EigenLoRAx demonstrates faster convergence compared to LoRA and VeRA. EigenLoRAx achieves a speedup of up to $1.5\times$ against LoRA and up to $2\times$ compared to PISSA.
This experiment was carried out on the CoLA task of the GLUE benchmark.}}
\label{fig:traininglosscola}
\end{figure*}
\subsection{GLUE Benchmark}~\label{sec:nlp}
Next, we evaluate EigenLoRAx on the General Language Understanding Evaluation (GLUE) benchmark~\citep{glue} datasets using the RoBERTa$_{base}$ model~\citep{roberta}. We use 6 different tasks: MRPC, SST-2, CoLA, QNLI, RTE and STS-B. Following the setup of VeRA, 
we omit time-intensive MNLI and QQP tasks, thus avoiding the use of MNLI initialization for MRPC, RTE, and STS-B tasks. In this setting, LoRAs are trained not on sub-datasets but on these different datasets representing a \textit{heterogeneous} domain setting, where the domain difference may be larger relative to the more domain-aligned setting in \cref{sec:img_class}. We follow the previous leave-one-out evaluation setup, where EigenLoRAx PCs are calculated using LoRAs of all but one task, and $\alpha$ is learnt for the left-out task. %
Refer to Appendix~\ref{appendix:glue} for all hyperparameters and implementation details.
\paragraph{Faster Convergence} Our results in \autoref{tab:glue_benchmark_results} show that EigenLoRAx ($K=32$) matches LoRA's performance with \textbf{100$\times$ fewer trainable parameters} and outperforms VeRA. EigenLoRAx extracts a useful principal subspace across diverse domains, enabling robust adaptation to new tasks. We also evaluate EigenLoRAx($^\text{init}$) weight initialization speed-up. Unlike PiSSA~\citep{meng_pissa_2024}, which initializes LoRA matrices with principal directions of pretrained weights, we randomly initialize weights within our extracted subspace. As shown in Figure~\ref{fig:traininglosscola}, EigenLoRAx converges faster than PiSSA and VeRA, and slightly faster than LoRA, highlighting the effectiveness of the principal subspace. VeRA's poorer performance may stem from suboptimal random initialization that fails to align with task-critical components. ELoRAx is also more efficient in terms of floating point operations for both forward and backward pass, as shown in~\cref{tab:flop_glue}. 
\paragraph{Low-Resource Scenario} 
To demonstrate the effectiveness of our subspace augmentation strategy~\cref{algo:eigenlora}, we conduct an experiment where EigenLoRAx is initialized with only 1–2 LoRAs. The results are presented in \cref{tab:glue_low}. We compare our method against augmenting EigenLoRAx with random components (EigenLoRAx+random) and using entirely random components (ELoRAx$^{\text{random}}$). As shown, our augmentation approach significantly outperforms random principal component selection. Interestingly, for MRPC, the base model's performance is retained. This suggests that the learned LoRA weights may not have influenced the base model, likely because they did not capture relevant information. While we do not provide theoretical guarantees for our principal component augmentation strategy—where randomly sampled vectors are iteratively orthogonalized to the existing EigenLoRAx principal vectors—we hypothesize that this targeted guidance helps prevent redundancy within the subspace. Consequently, it increases the likelihood of capturing the necessary task-relevant components.
\begin{table}[!h]
    \caption{Low-Resource GLUE Subset Results}
    \begin{center}
    \begin{sc}
    \begin{tabular}{lccc}
        \toprule
         & \# Param & MRPC & STS-B  \\
        \midrule
        ELoRAx$^{\text{RANDOM}}$ & 24K & 68.38 & -0.73 \\
        ELoRAx+rand & 24K  & 68.38 & 0.11  \\
        \rowcolor{gray!20} ELoRAx$^{\text{AUG}}$ & 24K & 83.09 & 85.28 \\
        \bottomrule
    \end{tabular}
    \label{tab:glue_low}
    \end{sc}
    \end{center}
\end{table}
\begin{table*}[!htb]
    \caption{\small{Results for Lots of LoRAs. We report the Rouge-L scores for each of the 5 tasks from the training set and 5 from the testing set. EigenLoRAx achieves on average \textbf{88\%} of LoRA's performance while requiring anywhere from \textbf{$12\times$} to \textbf{$95\times$} less parameters in a zero-shot setting.
}}  
    \begin{center}
    \begin{sc}
    \resizebox{\textwidth}{!}{%
    \begin{tabular}{lccccccccccccc}
        \hline
        & \# Trainable & \multirow{2}{*}{076} & \multirow{2}{*}{627} & \multirow{2}{*}{664} & \multirow{2}{*}{819} & \multirow{2}{*}{1631} & \multirow{2}{*}{039} & \multirow{2}{*}{290} & \multirow{2}{*}{391} & \multirow{2}{*}{442} & \multirow{2}{*}{1598}  & \multirow{2}{*}{Avg.} \\
        Method & Parameters & & & & & & & & & & &  \\
        \hline
        LoRA ($r=16$) & 9.4M  & 69.05 & 23.96 & 25 & 75 & 99.04 & 58.77 & 93.79 & 93.45 & 67.84 & 51.58 & 65.75  \\
        EigenLoRAx$^{\text{ZS}}$  & \textbf{98-786K} & 60.78 & 18.91 & 33.33 & 65.07 & 94.74 & 49.96 & 84.54 & 88.56 & 49.78 & 39.81 & 58.25  \\
        \rowcolor{gray!20} Performance Ratio & \textbf{ } & 0.88 & 0.79 & 1.33 & 0.87 & 0.96 & 0.79 & 0.90 & 0.95 & 0.73 & 0.77 & 0.88  \\
        \hline
    \end{tabular}%
    }
    \label{tab:lola}
    \end{sc}
    \end{center}
\end{table*}
\subsection{Lots of LoRAs}
\label{sec:lotsofloras}
Finally, we also tested our method in settings where a large number of adapters may be trained on significantly diverse domains. Lots of LoRAs \citep{brüelgabrielsson2024compressserveservingthousands} is a collection of over 500 adapters of the Mistral-7B-Instruct-v0.2 model \citep{jiang2023mistral7b}, trained on a variety of natural instruction tasks \citep{wang-etal-2022-super}. It represents the realistic setting where we directly use publicly available trained adapters, which may present significant diversity in terms of quality and task domain. As all adapters are accompanied with their respective training datasets, Lots of LoRAs is particularly useful in evaluating EigenLoRAx. The task presents significant diversity and a higher $K$ is necessary to represent this open domain. 

\paragraph{Setup} 
We split adapters randomly into two sets ($490, $5). EigenLoRAx PCs were calculated using the larger ``training" set and evaluations were done on the smaller ``test" set. We evaluated EigenLoRAx in a zero-shot setting (calculated using the already available adapter weights, no finetuning) . The results are shown in Table \ref{tab:lola} where we evaluate EigenLoRAx on the 5 tasks from the test set and also on 5 tasks from the training set to check for catastrophic forgetting or concept drift from scaling. The first 5 tasks are randomly sampled from the training set.
\begin{figure*}[!htb]
    \centering
    \begin{minipage}{0.49\textwidth}
        \centering
        \includegraphics[width=\textwidth]{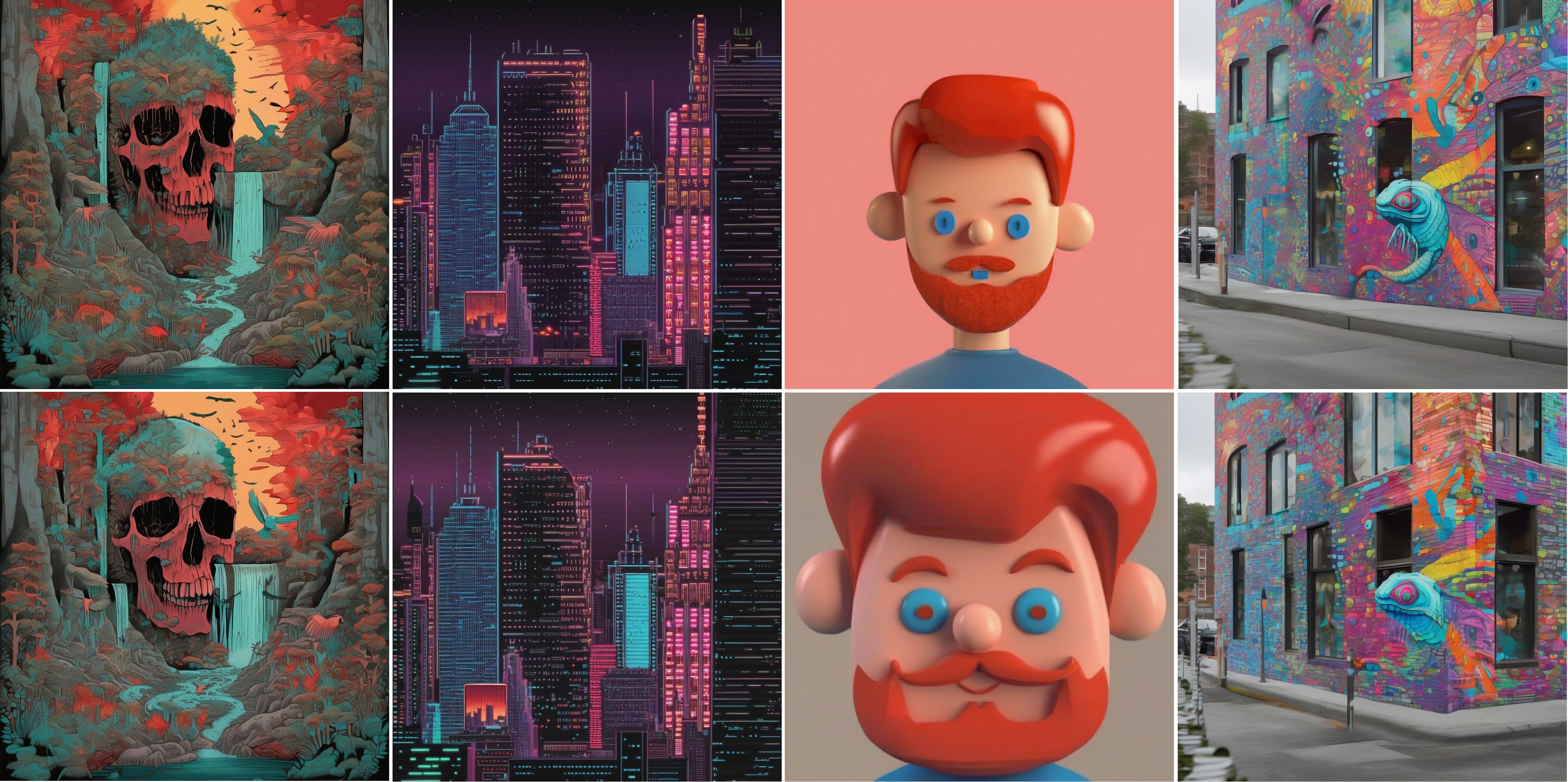} %
    \end{minipage}\hfill
    \begin{minipage}{0.49\textwidth}
        \centering
        \includegraphics[width=\textwidth]{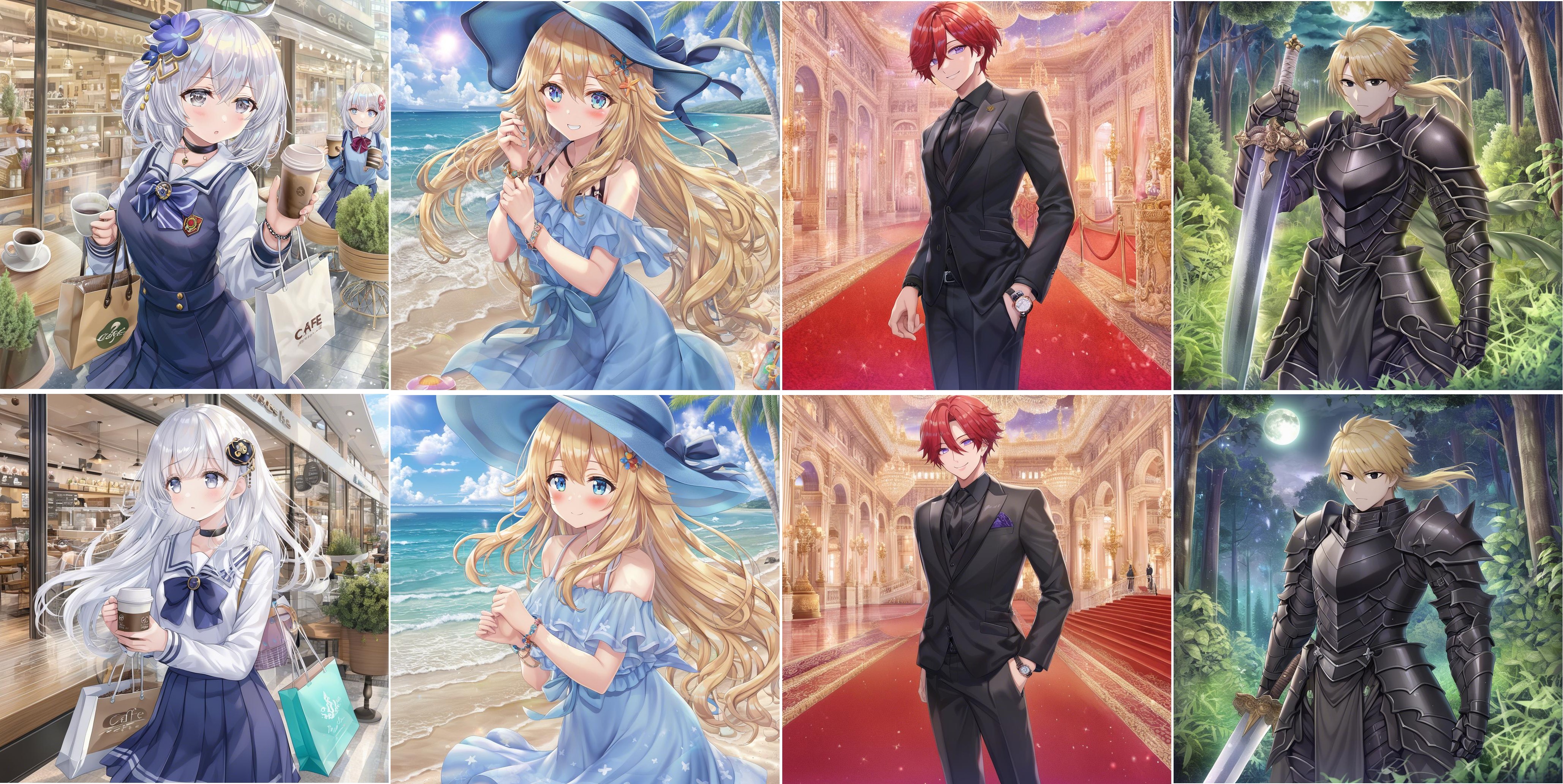} %
    \end{minipage}
    \caption{\small{LoRAs (top) vs. EigenLoRAx (bottom) in Text-to-Image generation. (Left) A single EigenLoRAx analytically reconstructs multiple LoRAs, significantly reducing memory (18$\times$ reduction) and compute costs. (Right) It efficiently learns new tasks with up to 100$\times$ fewer parameters than LoRA, maintaining similar visual quality. See \cref{appendix:diffusion} for more examples.}}
    \label{fig:diffusion_joint}
\end{figure*}
\textbf{EigenLoRAx nearly matches LoRA with $12 - 95\times$ fewer parameters}. EigenLoRAx recovers upto 88\% of LoRA's performance even in a zero-shot setting at such a large scale. The performance of EigenLoRAx can be improved by fine-tuning the EigenLoRAx adapters. In this setting we use randomized SVD in order to speed up the calculation of the PCs. We believe this leads to some degradation in performance as there randomized methods are approximations of the actual calculations. Performance can be further improved if better implementations of SVD which do not sacrifice accuracy for speed are used in calculating the Principal Components.
\subsection{Text-to-Image Image Generative Models}
\label{sec:diffusion}
We showcase EigenLoRAx's versatility on complex multimodal tasks like text-to-image generation, where LoRAs are extensively used to adapt models like Stable Diffusion to various styles and datasets. Despite thousands of LoRA adapters being available, most remain underutilized, occupying significant memory alongside their data. As adapter usage grows, a critical challenge is efficiently hosting multiple adapters for diverse tasks, especially on edge devices. Switching adapters during inference, often from CPU memory or disk, introduces latency that hinders real-time applications. EigenLoRAx tackles this by extracting a shared task-invariant subspace, significantly reducing in-memory parameters and enabling memory-efficient inference without compromising flexibility or performance.
EigenLoRAx can effectively replace pretrained adapters, drastically reducing storage requirements. To demonstrate this, we extracted $K=14$ principal components from $N=20$ Stable Diffusion-XL~\citep{sdxl} LoRA adapters (rank $r=32$) from the HuggingFace diffusers library~\citep{von-platen-etal-2022-diffusers}. Using $\alpha \in \mathbb{R}^{r \times K}$, we analytically reconstructed the original LoRA weights within the extracted principal subspace. For image generation, we used 30 denoising steps with a fixed seed of 0. Results and comparisons are shown in \cref{fig:diffusion_joint}. This approach reduces storage requirements for all adapters from 4.6GB to just 261MB, achieving an \textbf{18$\times$ reduction in low-rank parameters stored in memory}. By enabling a large number of adapters to reside in VRAM simultaneously, EigenLoRAx eliminates I/O bottlenecks, significantly improving memory efficiency for real-time applications.
\begin{figure*}[!hbt]
\begin{center}
\includegraphics[width=0.9\textwidth]{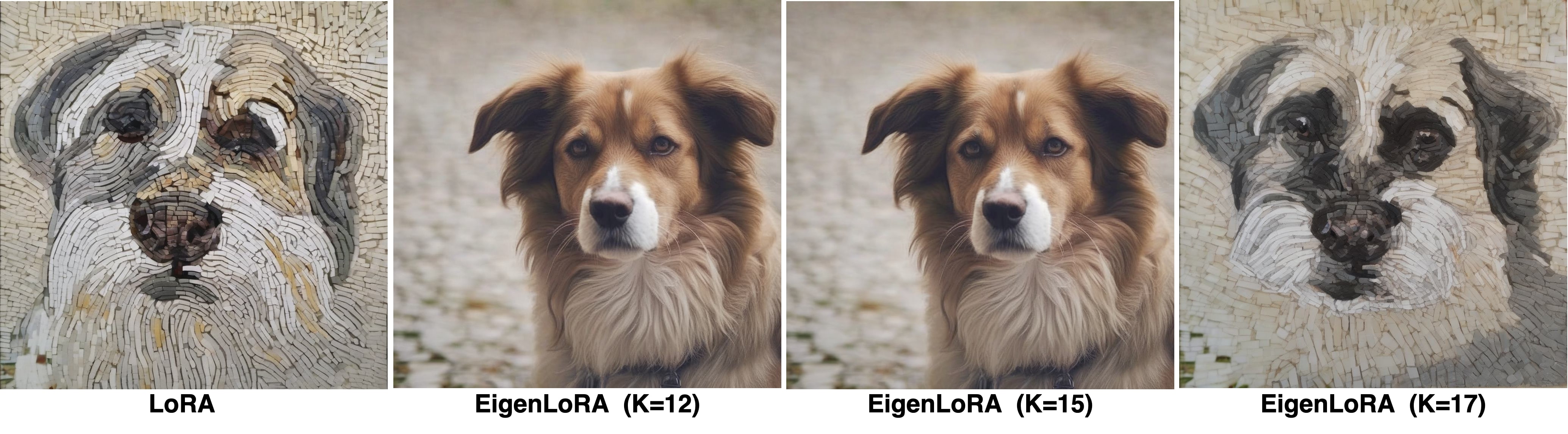}
\end{center}
\caption{Failure Case: EigenLoRAx may fail if an important component is missing from the initialized subspace i.e. the shared subspace is incomplete, which may happen due to inadequacy in the number of initial adapters or due to the majority of the adapters being of bad quality. E.g., the model may have lost the essential "mosaic" property when generating an image for the prompt: "mosaic picture of a dog."}
\label{fig:failurediffusion}
\end{figure*}
\paragraph{Failure Cases and Limitations} Despite its advantages, EigenLoRAx has limitations. Figure~\ref{fig:failurediffusion} shows a failure case where the method fails to capture a key property of the desired image. While tasks may share a principal subspace, missing critical orthogonal components can degrade performance, especially if they were absent in the pretrained LoRAs used for extraction or if the chosen top $K$ components were suboptimal. In the latter case, empirical analysis of hyperparameters (\cref{sec:ablation}) can guide optimal $K$ selection. Additionally, our subspace augmentation method (\cref{tab:glue_low}) helps by iteratively sampling and orthogonalizing more components to recover missing subspace elements. A simple extension can further mitigate this issue by allowing a small number of rank-1 weights to be trainable outside the subspace. Another key limitation (\cref{sec:lotsofloras}) is the computational cost and instability of processing a large number of initial LoRAs. A continual learning approach building on our method could address this. Finally, our experiments did not explore layer-wise or weight matrix-level optimizations; we tested different $K$ values but kept them fixed across layers and for both A and B matrices. Additional failure cases are discussed in \cref{ssec:failure_append}.

%% file: 8_appendix.tex
\section{Appendix}\label{sec:appendix}

\subsection{Experiments}\label{appendix:hyperparameters}
For VeRA, LoRA and PiSSA, we experimented with a range of learning rates, from higher to lower, along with three different scheduling approaches: ReduceLRonPlateau, Linear, and Cosine. The hyperparameters that yielded the best average performance were selected for further experimentation. The observed discrepancies with EigenLoRAx hyperparameters are attributable to these methodological choices. Comprehensive hyperparameter tuning for EigenLoRAx was not pursued extensively, as the initially selected hyperparameters, notably a high learning rate paired with ReduceLRonPlateau or Linear, demonstrated satisfactory performance, thereby conserving computational resources.

\subsubsection{Image Classification}
\paragraph{Trainable parameters for EigenLoRAx}
The base model is vit-base-patch16-224. The following are the trainable parameters in ViT~\citep{vision_transformer} that are trained for EigenLoRAx. We ignore the last linear layer for simplicity since it is trained for all models and baselines and is constant. The loading parameter has the shape of $[\text{number of EigenLoRAx PC} , 1]$ (we only have $2$ in each EigenLoRAx PC for this experiment). Therefore, the total number of trainable parameters (for the number of components$=2$) is $12\text{ (layers) }\times4\text{ (set of parameters per layers) }\times2\text{ (number of trainable parameter per coefficient) } = 96$ trainable parameters.

\paragraph{Hyperparameters} LoRA~\citep{hu2021lora} and VeRA~\citep{kopiczko_vera_2023} implementations are taken from the HuggingFace PEFT~\citep{peft} library with hyperparameters of the default method. For Food101~\citep{food101} experiment, we randomly remove 1 class for ease of compute. Experimental hyperparameters are reported in \autoref{tab:appendix_classn_lora_hyp} and \autoref{tab:appendix_classn_leora_hyp}.
\begin{table}[h]
  \centering
   \caption{Hyperparameters for LoRA~\citep{hu2021lora} and VeRA~\citep{kopiczko_vera_2023} for the Image Classification Experiment}
  \begin{tabular}{cccc}
    \toprule
    & \textbf{CIFAR100} & \textbf{Flowers102} & \textbf{Food101} \\
    \midrule
    Learning Rate            & $1\mathrm{e}{-4}$
 & $1\mathrm{e}{-4}$ & $1\mathrm{e}{-4}$  \\ 
    Weight Decay            & 0.1      & 0.1      & 0.1          \\ 
    Warmup ratio  & 0.06     & 0.06     & 0.06      \\ 
    Epochs        & 10       & 10       & 10             \\ 
    Number of Subsets          & 5       & 6        & 5        \\ 
    Categories/Subset          & 20      & 17        & 20     \\
    Seed          & 42        & 42        & 42            \\ 
    Batch Size          & 128       & 64        & 128      \\ 
    \bottomrule
  \end{tabular}
  \label{tab:appendix_classn_lora_hyp}
\end{table}

\begin{table}[h]
  \centering
    \caption{Hyperparameters for EigenLoRAx for the Image Classification Experiment}
  \begin{tabular}{ccccc}
    \toprule
    & \textbf{CIFAR100} & \textbf{Flowers102} & \textbf{Food101}  \\
    \midrule
    Learning Rate            & $1\mathrm{e}{-2}$
 & $1\mathrm{e}{-2}$ & $1\mathrm{e}{-2}$ \\ 
    Weight Decay            & 0.1      & 0.1      & 0.1        \\ 
    Warmup ratio  & 0.06     & 0.06     & 0.06      \\ 
    Epochs        & 10       & 10       & 10            \\ 
    Number of Subsets          & 5       & 6        & 5   \\ 
    Categories/Subset          & 20      & 17        & 20  \\
    Seed          & 42        & 42        & 42        \\ 
    Batch Size          & 128       & 64        & 128     \\ 
    \bottomrule
  \end{tabular}
  \label{tab:appendix_classn_leora_hyp}
\end{table}

\paragraph{Experimental Results} The experiments were conducted $5$ times utilizing randomly generated dataset splits. The mean accuracy values are reported in \autoref{tab:vision_models}. Empirical analysis indicates that without control and annealing of learning rates, the loss for both LoRA and VeRA may diverge or plateau, particularly with high learning rates. Even with the lower learning rate, Full training or LoRA can overfit to the training data without proper regularization. In contrast, no such instability was observed during EigenLoRAx training, where a relatively higher learning rate proved advantageous for rapid convergence. 

\begin{table}[h]
    \centering
    \caption{Image Classification Accuracy results on CIFAR100~\citep{cifar100}}
    \begin{tabular}{llcccccc}
        \toprule
         & \textbf{Trainable} & &  &  &  &  & \\
        \textbf{Model} & \textbf{Params} & \textbf{subset1} & \textbf{subset2} & \textbf{subset3} & \textbf{subset4} & \textbf{subset5} & \textbf{Avg.} \\ 
        \midrule
        \textbf{FT}             & 86389248             & 98.8                        & 97.95                       & 95.55                       & 96.05                       & 96.3                        & 96.93                       \\
        \textbf{LoRA ($r=1$)}   & 36864                & 97.6                        & 93.95                       & 93.75                       & 91.75                       & 85.2                        & 92.45                       \\
        \textbf{LoRA ($r=4$)}   & 147456               & 98.15                       & 95.2                        & 93.5                        & 92.85                       & 89.25                       & 93.79                       \\
        \textbf{VeRA ($r=2$)}   & 18480                & 93.65                       & 89.7                        & 89.5                        & 89.95                       & 91.55                       & 90.87                       \\
        \textbf{EigenLoRAx ($K=2$)} & 96                   & 97.25                       & 95.05                       & 94.55                       & 93                          & 94.15                       & 94.8                        \\
            \bottomrule
    \end{tabular}
    \label{tab:appendix_cifar100}
\end{table}

\begin{table}[h]
    \centering
    \caption{Image Classification Accuracy results on Food101~\citep{food101}}
    \begin{tabular}{llcccccc}
        \toprule
         & \textbf{Trainable} & &  &  &  &  & \\
        \textbf{Model} & \textbf{Params} & \textbf{subset1} & \textbf{subset2} & \textbf{subset3} & \textbf{subset4} & \textbf{subset5} & \textbf{Avg.} \\ 
        \midrule
        \textbf{FT}               & 86389248             & 98.64                       & 97                          & 97.36                       & 94.28                       & 95.92                       & 96.64                       \\
\textbf{LoRA ($r=1$)}   & 36864                & 93.36                      & 88.44                       & 94.28                       & 89.4                        & 89.9                        & 91.076                      \\
\textbf{LoRA ($r=4$)}   & 147456               & 98.2                        & 96.96                       & 96.08                       & 92.88                       & 94.52                       & 95.728                      \\
\textbf{VeRA ($r=2$)}   & 18480                & 91.22                       & 88.42                       & 94.42                       & 91.88                       & 92.82                       & 91.752                      \\
\textbf{EigenLoRAx ($K=2$)} & 96                   & 97.24                       & 95.96                       & 96                          & 91.88                       & 94.6                        & 95.136                      \\
            \bottomrule
    \end{tabular}
    \label{tab:appendix_food101}
\end{table}

\begin{table}[h]
    \centering
    \caption{Image Classification Accuracy results on Flowers102~\citep{flowers102}}
    \begin{tabular}{lccccccc}
        \toprule
        \textbf{Model}  & \textbf{subset1} & \textbf{subset2} & \textbf{subset3} & \textbf{subset4} & \textbf{subset5} & \textbf{subset6} & \textbf{Avg.} \\ 
        \midrule
        \textbf{FT}                          & 99.7                        & 99.3                        & 98.01                       & 98.22                       & 99.7                        & 98.01                       & 98.82          \\
\textbf{LoRA ($r=1$)}                   & 85.9                        & 88.47                       & 92.69                       & 91.02                       & 91.7                        & 91.01                       & 90.13          \\
\textbf{LoRA ($r=4$)}                  & 96.23                       & 92.76                       & 97.22                       & 95.01                       & 98.24                       & 90.73                       & 95.03         \\
\textbf{VeRA ($r=2$)}                & 99.2                        & 95.4                        & 97.7                        & 94.7                        & 90.9                        & 95                          & 95.48        \\
\textbf{EigenLoRAx ($K=2$)}                    & 99.686                      & 97.905                      & 97.689                      & 98.291                      & 99.344                      & 97.718                      & 98.43         \\
            \bottomrule
    \end{tabular}
    \label{tab:appendix_flowers102}
\end{table}
\FloatBarrier

\subsection{Natural Language Processing - GLUE benchmark}~\label{appendix:glue}

\paragraph{Hyperparameters} LoRA~\citep{hu2021lora}, VeRA~\citep{kopiczko_vera_2023} and PISSA~\citep{meng_pissa_2024} implementations are taken from the HuggingFace PEFT~\citep{peft} library. Refer to \autoref{tab:appendix_glue_lora_hyp} and \autoref{tab:appendix_glue_elora_hyp} for hyperparameter details. For LoRA~\citep{hu2021lora}, we use the ranks $\in \{8,16\}$. For VeRA~\citep{kopiczko_vera_2023}, we use rank$=256$, and for EigenLoRAx, we use $K\in\{16, 32\}$ and $r=8$. Here, $r$ refers to the dimensionality of the trainable coefficients and not the rank. For both PISSA~\citep{meng_pissa_2024} and LoRA, all the parameters of the low rank matrix are trainable. For the EigenLoRAx initialization experiment, we train both the components and coefficients for a fair comparison with PISSA. In practice, however, we do not need to do so - we can tune only the sparse coefficients and after the loss converges, finetune the components for a few training steps.

\begin{table}[h]
  \centering
   \caption{Hyperparameters for LoRA~\citep{hu2021lora}, VeRA~\citep{kopiczko_vera_2023} and PiSSA~\citep{meng_pissa_2024} for the GLUE benchmark.~\citep{glue}}
  \begin{tabular}{ccccccc}
    \toprule
    & CoLA & MRPC & QNLI & RTE & SST-2 & STSB \\
    \midrule
    Learning Rate            & $4\mathrm{e}{-4}$
 & $4\mathrm{e}{-4}$ & $4\mathrm{e}{-4}$ & $5\mathrm{e}{-4}$ & $5\mathrm{e}{-4}$ & $4\mathrm{e}{-4}$ \\ 
    Weight Decay            & 0.1      & 0.1      & 0.1      & 0.1      & 0.1      & 0.1      \\ 
    Warmup ratio  & 0.06     & 0.06     & 0.06     & 0.06     & 0.06     & 0.06     \\ 
    Epochs        & 80       & 30       & 25       & 80       & 60       & 40       \\ 
    Scheduler     & Linear   & Linear   & Linear   & Linear   & Linear   & Linear   \\
    Seed          & 0        & 0        & 0        & 0        & 0        & 0        \\ 
    Batch Size          & 64       & 64        & 64        & 64        & 64        & 64\\ 
    \bottomrule
  \end{tabular}
  \label{tab:appendix_glue_lora_hyp}
\end{table}

\begin{table}[h]
  \centering
    \caption{Hyperparameters for EigenLoRAx for the GLUE benchmark.~\citep{glue}.\\ (RLrP - ReduceLRonPlateau)} 
  \begin{tabular}{ccccccc}
    \toprule
    & CoLA & MRPC & QNLI & RTE & SST-2 & STSB \\
    \midrule
    Learning Rate            & $4\mathrm{e}{-3}$
 & $4\mathrm{e}{-3}$ & $4\mathrm{e}{-3}$ & $5\mathrm{e}{-3}$ & $5\mathrm{e}{-3}$ & $4\mathrm{e}{-3}$ \\ 
    Weight Decay            & 0.1      & 0.1      & 0.1      & 0.1      & 0.1      & 0.1      \\ 
    Warmup ratio  & 0.06     & 0.06     & 0.06     & 0.06     & 0.06     & 0.06     \\ 
    Epochs        & 80       & 30       & 25       & 80       & 60       & 40       \\ 
    Scheduler     & RLrP & RLrP   & RLrP   & RLrP & RLrP   & RLrP   \\
    Seed          & 0        & 0        & 0        & 0        & 0        & 0        \\
    Batch Size          & 64       & 64        & 64        & 64        & 64        & 64\\ 
    \bottomrule
  \end{tabular}
  \label{tab:appendix_glue_elora_hyp}
\end{table}
\FloatBarrier

\FloatBarrier
\subsection{Text-to-Image Generation (Stable Diffusion Models)}\label{appendix:diffusion}
\begin{figure}[htb]
\begin{center}
\includegraphics[width=0.8\textwidth]{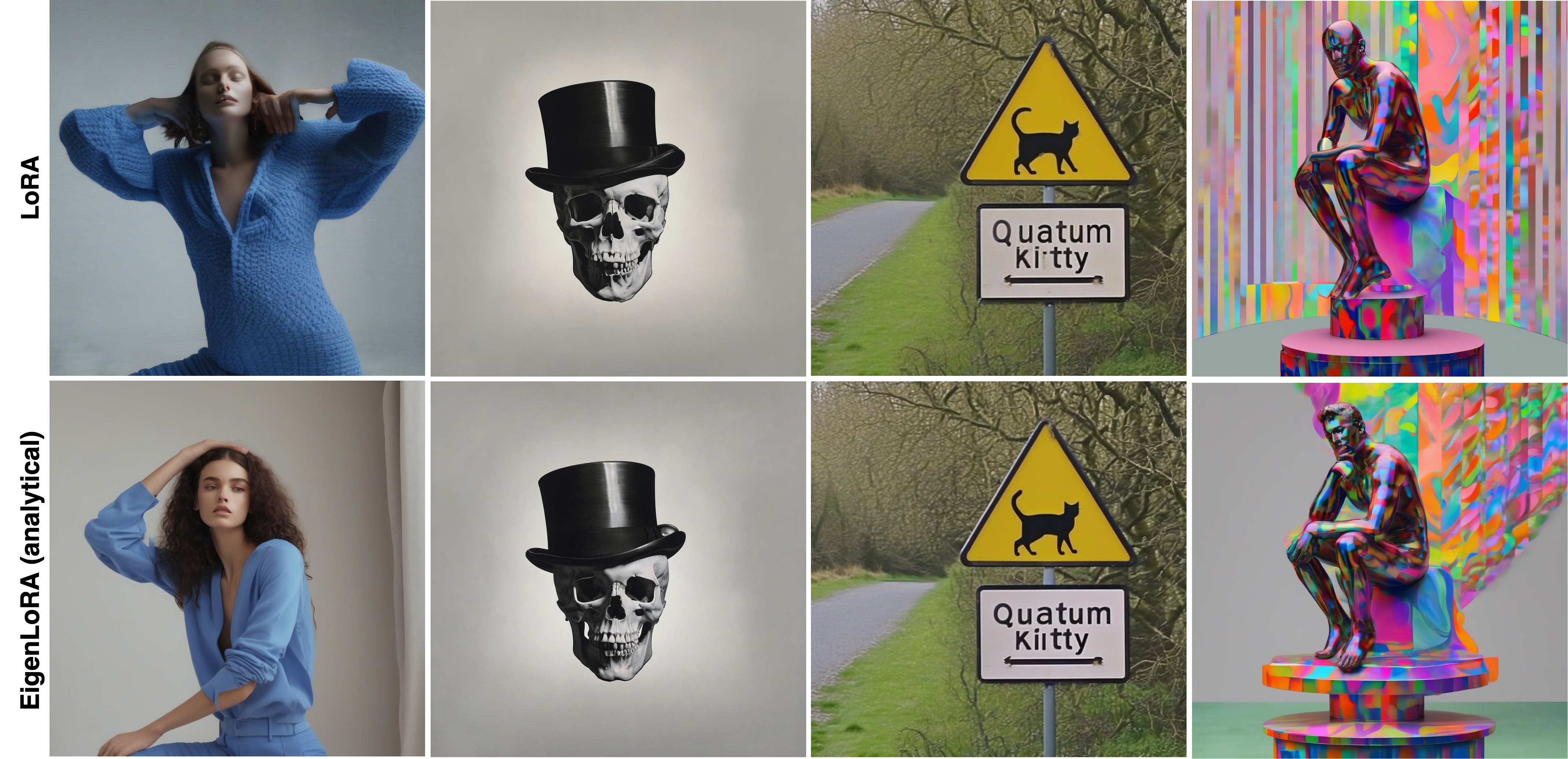}
\end{center}
\caption{(Part 1) A single EigenLoRAx (identical components, varying loadings) was employed to produce these images utilizing the Stable Diffusion-XL~\cite{sdxl} model. A comparison between our results and those obtained from multiple LoRAs does not show a noticeable degradation in visual quality.}
\label{fig:diffusion3}
\end{figure}

\begin{figure}[htb]
\begin{center}
\includegraphics[width=0.8\textwidth]{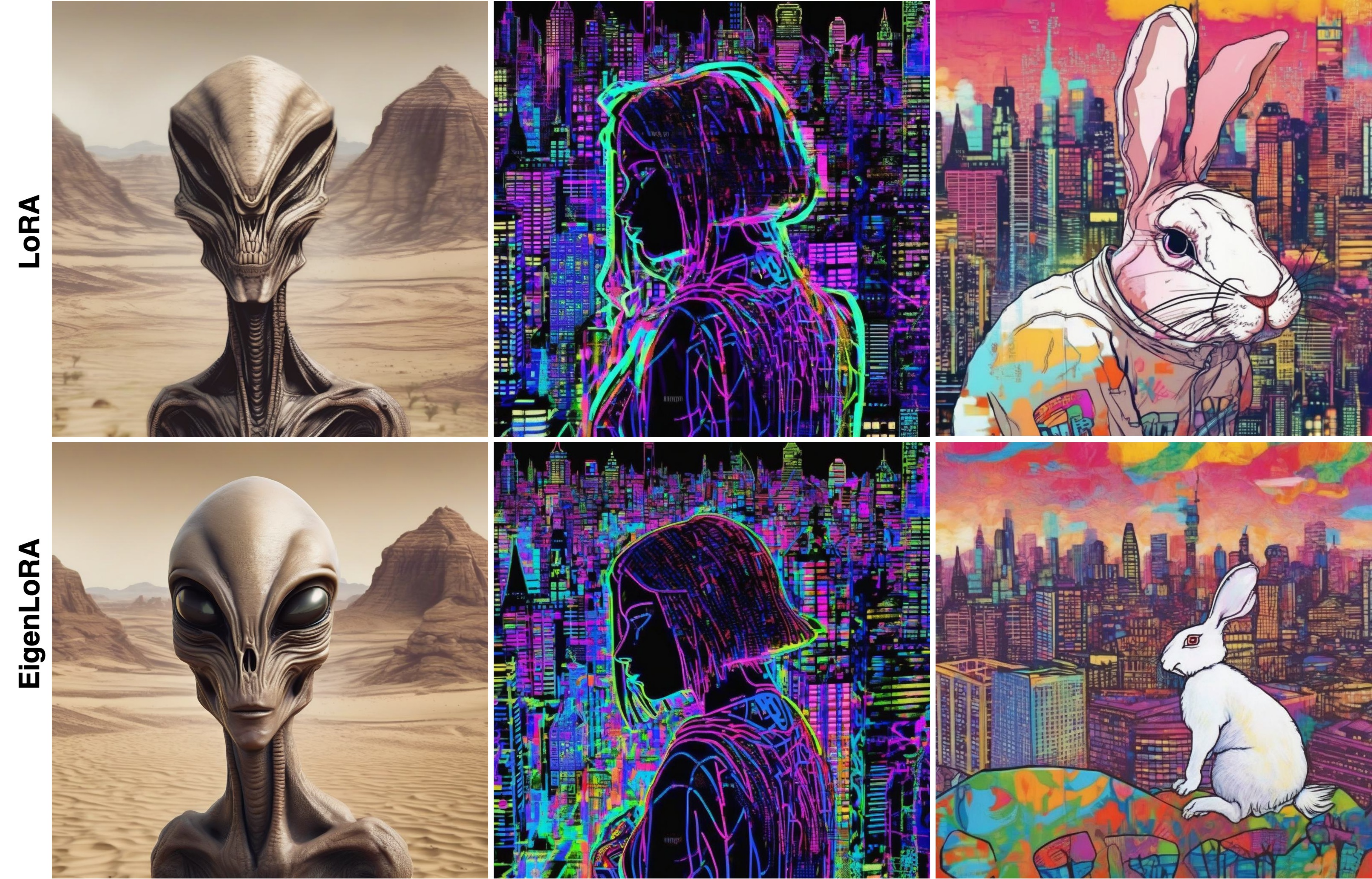}
\end{center}
\caption{(Part 2) A single EigenLoRAx (identical components, varying loadings) was employed to produce these images utilizing the Stable Diffusion-XL~\cite{sdxl} model. A comparison between our results and those obtained from multiple LoRAs demonstrates no noticeable degradation in visual quality. }
\label{fig:diffusion2}
\end{figure}

\autoref{fig:diffusion3} and \autoref{fig:diffusion2} show more examples of a text-to-image stable diffusion model finetuned using EigenLoRAx. Note that not only there is no publicly available code for VeRA that allows its usage in complex text-to-image generation tasks, but our VeRA implementation also did not work well in this task.

\begin{figure}[htb]
\begin{center}
\includegraphics[width=0.8\textwidth]{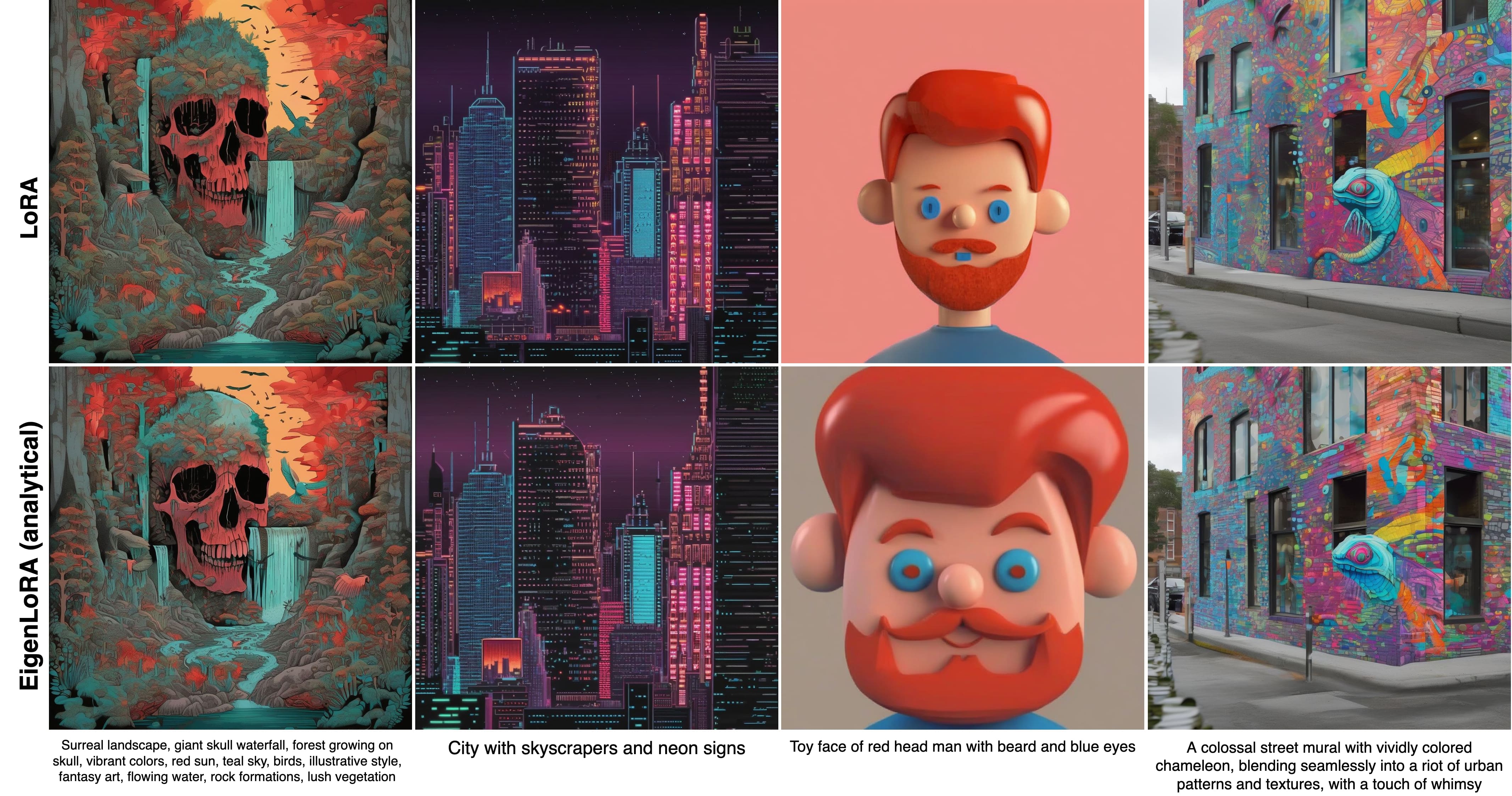}
\end{center}
\caption{
\small{Analytical reconstruction of LoRAs using EigenLoRAx which shows no degradation in relative visual quality. 
See Appendix~\ref{appendix:diffusion} for more examples.}}
\label{fig:diffusion}
\end{figure}

\begin{figure}[htb]
\begin{center}
\includegraphics[width=0.9\textwidth]{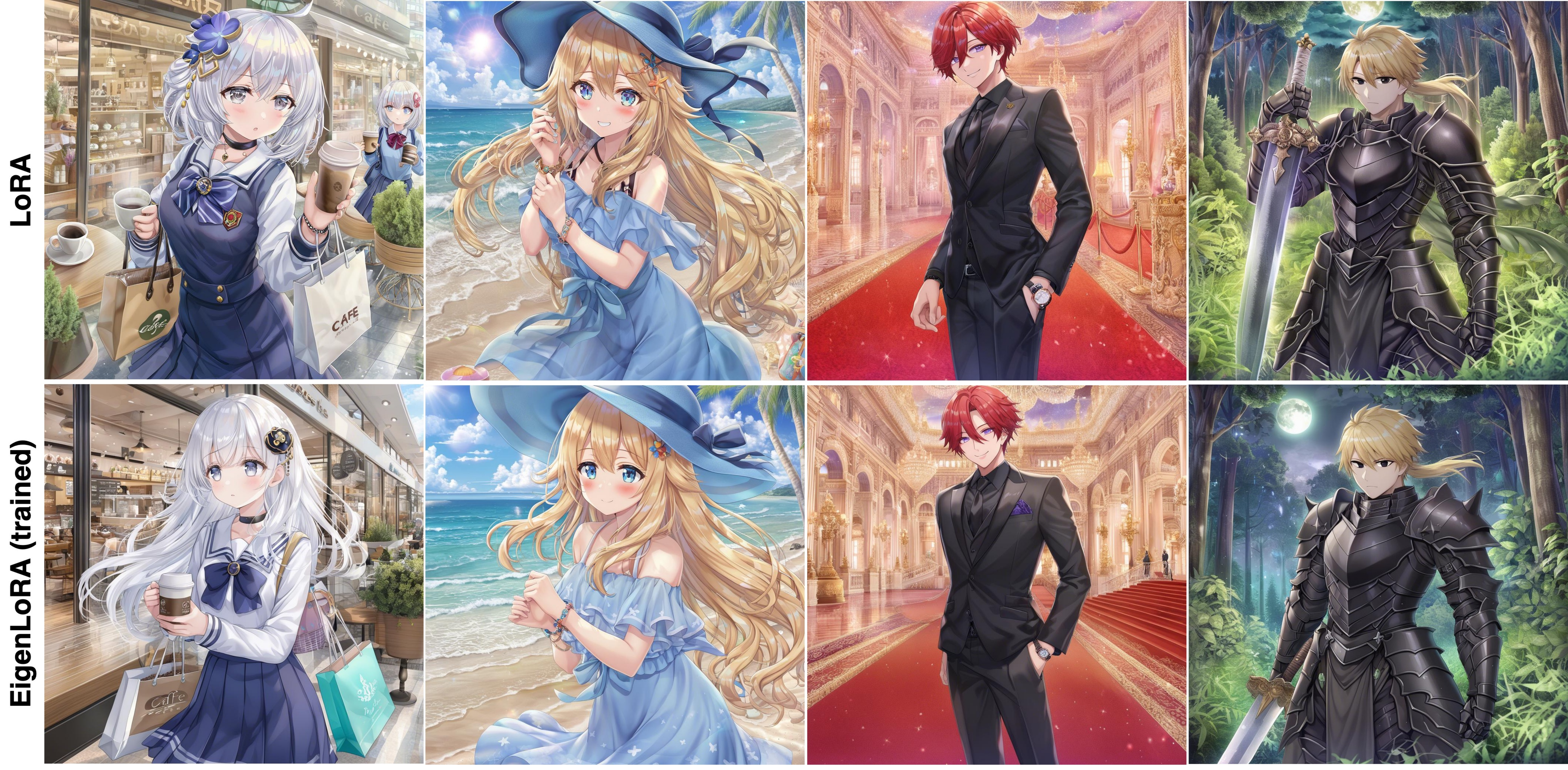}
\end{center}
\caption{\small{
Comparison of generated images by LoRA and EigenLoRAx trained on Torino Aqua anime style images. 
For EigenLoRAx, we utilized 12 components with only trainable coefficients to finetune the base model.
}}
\label{fig:traindiffusion}
\end{figure}
\FloatBarrier

\subsection{Additional Experiments}
Furthermore, we also performed a 3D object pose estimation~\citep{wang2020NeMo, Kaushik2024SourceFreeAI} finetuning experiment using a modified Resnet-101. The task of 3D object pose estimation involves the prediction of three rotation parameters (azimuth, elevation, in-plane rotation) of an object relative to the camera. The pose estimation error between the predicted rotation matrix and the ground truth rotation matrix is given as
$\Delta (R_{pred}, R_{gt}) = \frac{\vert\vert \log_b (R_{pred}^{\intercal} R_{gt}) \vert\vert_F}{\sqrt{2}} $
We show the results for the $\frac{\pi}{6}$ accuracy threshold for this experiment.

\begin{table}[!ht]
    \centering
        \caption{3D object pose estimation accuracy ($\frac{\pi}{6}$ threshold)}
    \begin{tabular}{lcccccccc}
    \toprule
        Method & Param & Airplane & Motorbike & Boat & Bottle & Bus & Car & Average \\ 
        \midrule
        LoRA ($r=16$) & 215K & 79.9 & 80.1 & 71.5 & 89.8 & 90.1 & 96.6 & 84.67 \\ 
        VeRA ($r=256$) & 40K & 68.4 & 72.4 & 64.3 & 88.4 & 87.2 & 94.4 & 79.18 \\ 
        EigenLoRAx ($K=2$) & 16K & 81.4 & 80.0 & 71.4 & 90 & 92.3 & 97.5 & 85.43 \\ 
        \bottomrule
    \end{tabular}
\end{table}
\FloatBarrier
\section{Method Analysis and Ablation}
\label{sec:ablation}
\begin{figure}[h]
  \begin{center}
    \includegraphics[width=.8\textwidth]{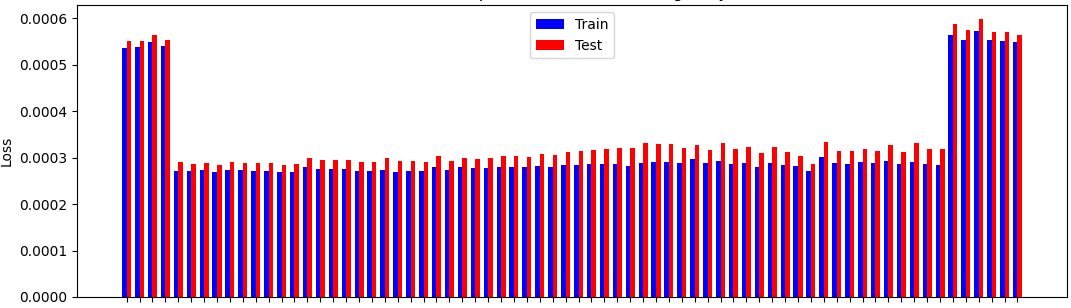}
  \end{center}
  \caption{\small{Average reconstruction error between EigenLoRAx and a set of LoRAs for all UNet layers in a stable diffusion model.}}
  \label{fig:reconserror}
\end{figure}

Through a rigorous comparative analysis of EigenLoRAxs and their target LoRAs, we identified that the most pronounced reconstruction discrepancies manifest in the initial and terminal layers of the neural network, as depicted in~\autoref{fig:reconserror}. Allowing the EigenLoRAx PCs in these layers to undergo fine-tuning alongwith the coefficients can alleviate failure scenarios, thereby alleviating the need for comprehensive model fine-tuning.

\subsection{How to Choose $K$ Principal Components and $r$ for EigenLoRAx}
\begin{figure}[!htb]
    \centering
    \begin{minipage}{0.49\textwidth}
        \centering
        \includegraphics[width=\textwidth]{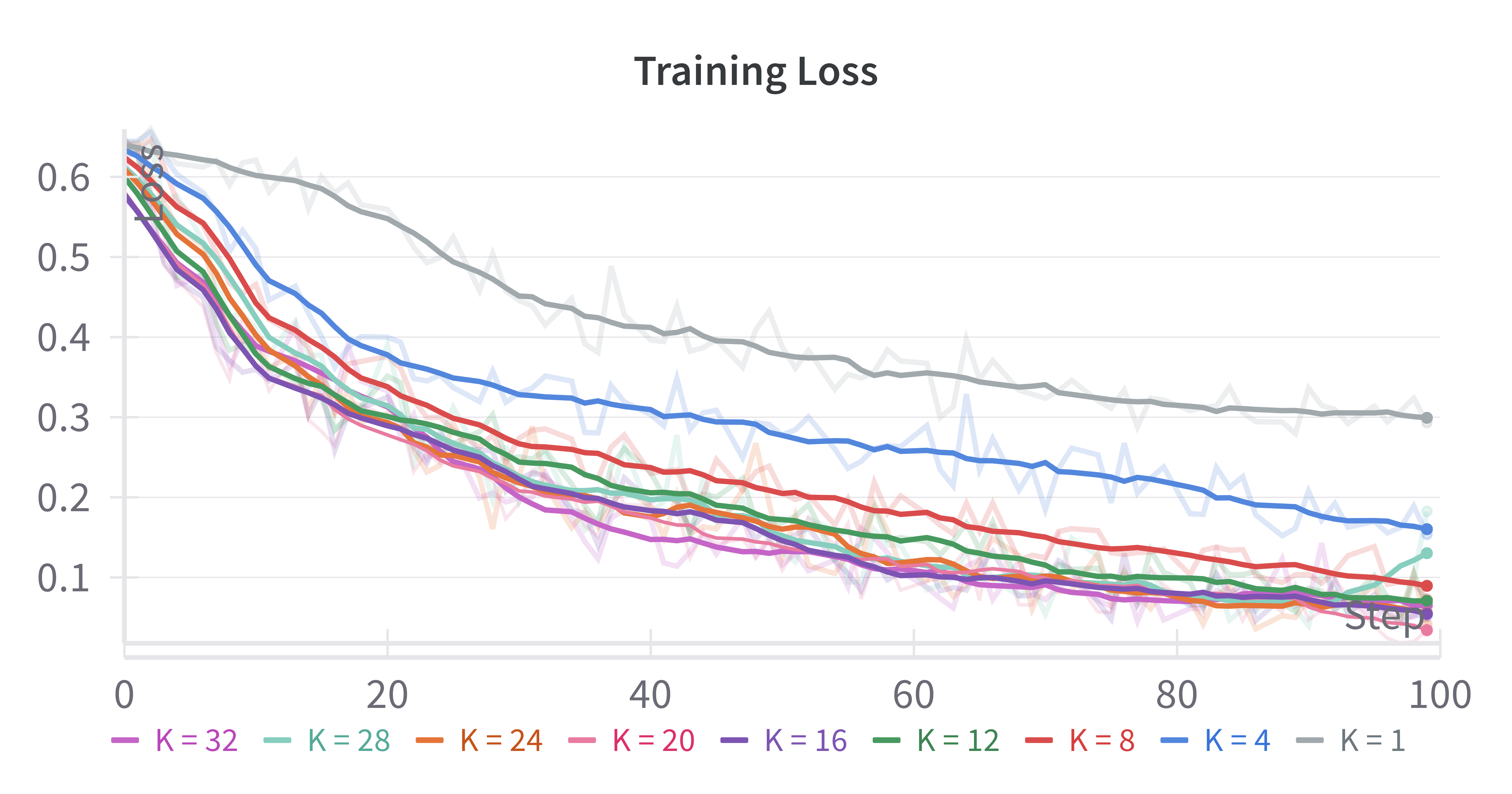} %
        \caption{Training Loss Convergence for different numbers of EigenLoRAx PCs}
        \label{fig:mrpc_loss}
    \end{minipage}\hfill
    \begin{minipage}{0.49\textwidth}
        \centering
        \includegraphics[width=\textwidth]{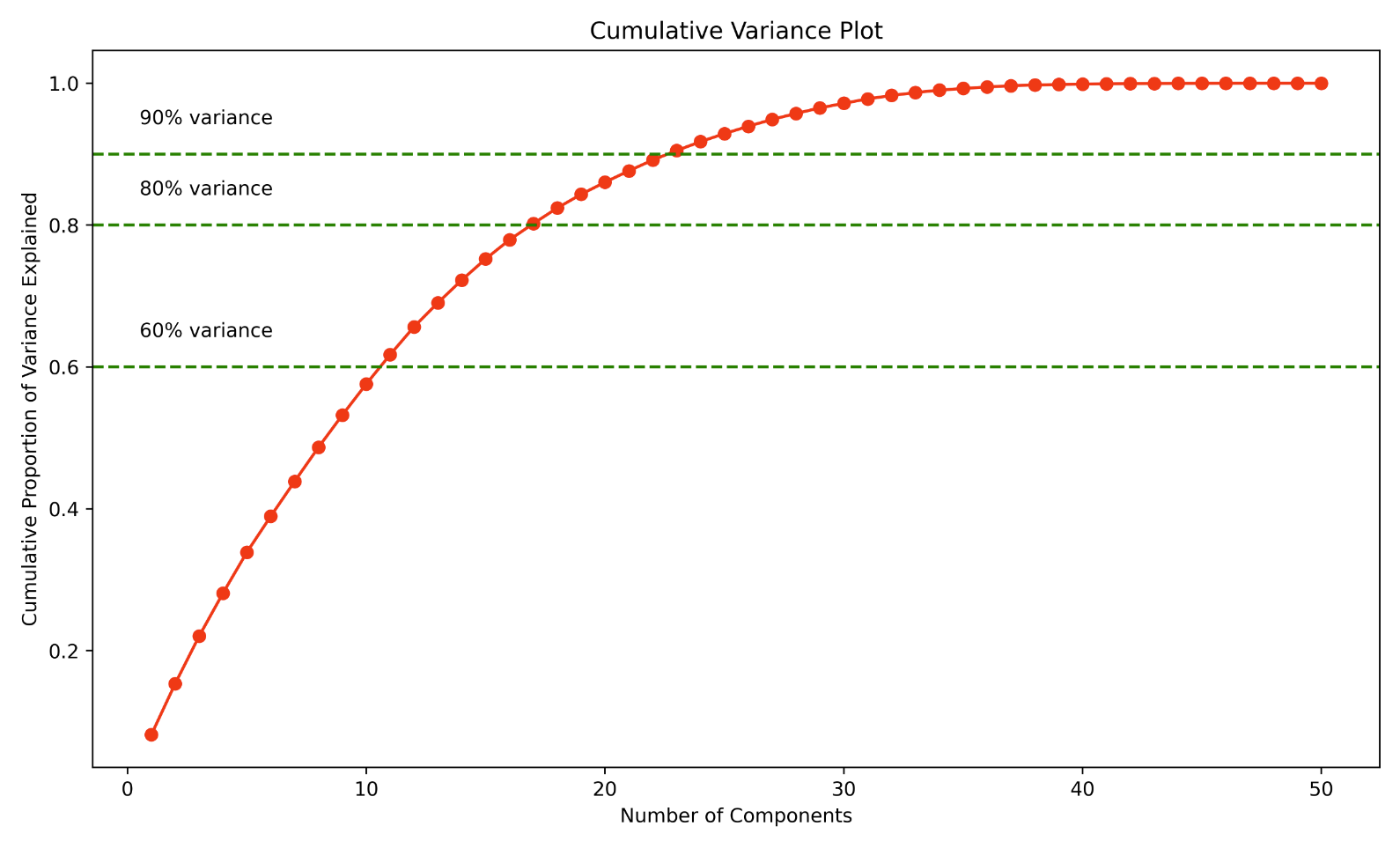} %
        \caption{Explained Variance for increasing number of PCs}
    \label{fig:mrpc_var}
    \end{minipage}
    \caption{Ablation of Number of EigenLoRAx Principal Components}
    \label{fig:mrpc_K}
\end{figure}

\begin{figure}[h]
  \begin{center}
    \includegraphics[width=\textwidth]{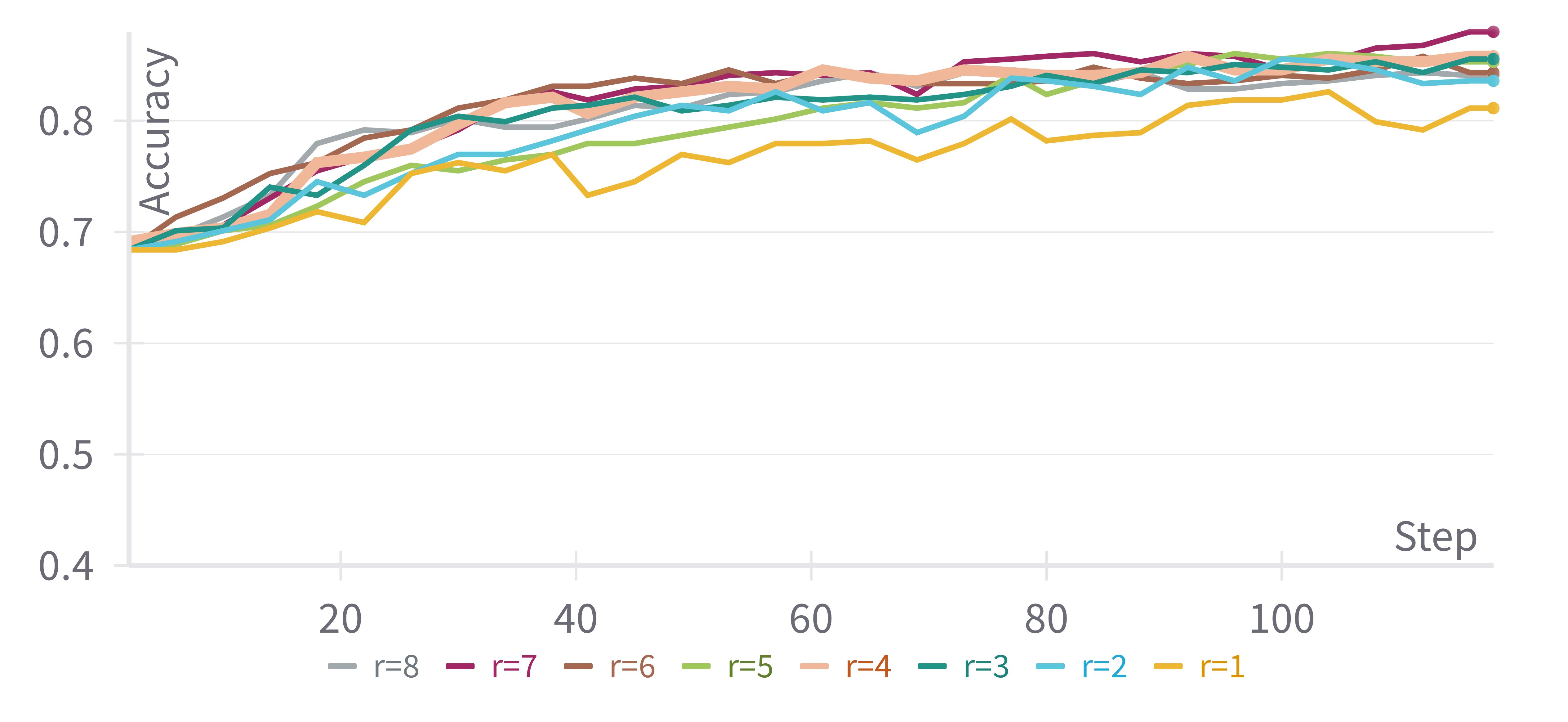}
  \end{center}
  \caption{Ablation for the EigenLoRAx's $r$ hyperparameter. This experiment was done for the MRPC task in the GLUE benchmark.}
  \label{fig:r_ablate}
\end{figure}

We perform an ablation study on the selection of EigenLoRAx principal components ($K$). Our analysis concentrates on one experiment as shown in ~\autoref{fig:mrpc_K}, specifically pertaining to the MRPC task within the GLUE~\citep{glue} benchmark. The analysis in \autoref{fig:mrpc_loss} shows the training loss in relation to increasing number of EigenLoRAx principal components $K$, as well as the explained variance of the LoRAs used to initialize the EigenLoRAx in \autoref{fig:mrpc_var}. We find, empirically, that choosing EigenLoRAx PCs for the explained variance of $50-80\%$ of the LoRAs used to initialize EigenLoRAx is sufficient for a robust initialization. This is shown in \cref{fig:mrpc_var} where we choose $K=8$ which roughly corresponds to the explained variance of $55-60\%$. We further ablate this choice in \cref{fig:mrpc_loss}, where although substantial improvements are evident up to $K = 8$, an increase in the number of $K$ thereafter yields only marginal gains, demonstrating diminishing returns as the number of components increases. The parameter $r$ in EigenLoRAx does not equate the \textit{rank} parameter in LoRA and its variants. It reflects the dimensionality of the EigenLoRAx coefficients. Although $r=1$ works well, we observe slight performance improvements as we increase this value as shown in \cref{fig:r_ablate}. Increasing this value corresponds to a small amount of parameter increase. We observe no finetuning instability by changing this value and recommend that it can be set to anywhere between 1 and the rank of the LoRAs used to initialize EigenLoRAx.
\subsection{Failure Cases}
\label{ssec:failure_append}
\autoref{fig:failurediffusion} illustrates a potential failure case of EigenLoRAx, where the incorrect number of principal components (PCs) was selected. In this instance, the "mosaic style" information was excluded from the principal subspace identified by EigenLoRAx due to an insufficient number of PCs. However, this issue can be resolved by selecting a larger number of PCs, as the extended principal subspace contains the necessary information for the task.

Another hypothetical failure scenario arises if the domain gap between the low-rank adapters used to initialize EigenLoRAx and the downstream task is significantly large. Although we do not observe such a case in our experiments, it is plausible that under such conditions, EigenLoRAx might underperform. This issue could potentially be mitigated by allowing only a subset of PCs to remain trainable, enabling the model to adapt more effectively to the target domain.

A further observed limitation of EigenLoRAx occurs in complex tasks like Text-to-Image generation, which may extend to other tasks as well. If the majority of LoRAs used to initialize EigenLoRAx encode biases (e.g., related to gender, race, or context), these biases tend to propagate into EigenLoRAx outputs. While such biases are a common issue in deep learning models trained using stochastic gradient descent or similar methods, addressing them remains a critical area of future work. We consider this an important avenue for improvement and discuss the broader implications in \cref{sec:impact}.

\subsection{Impact of LoRA adapter quality on EigenLoRAx PC initialization}

To evaluate EigenLoRAx’s robustness to adapter quality and its resistance to noise, we conducted an ablation study on a subset of tasks of the NLU experiment specified in Section~\ref{sec:nlp}. Specifically, we generated EigenLoRAx adapters using LoRA matrices with varying levels of random noise added. The results are shown in \autoref{tab:noise}

\begin{table}[!h]
    \centering
        \caption{EigenLoRAx performance on subset of GLUE task using noisy LoRA adapters for initialization}
    \begin{tabular}{cccccc}
    \toprule
        Noise Level & CoLA & MRPC & RTE & STS-B & Avg \\ 
        \midrule
        5\% & 60.51 & 85.45 & 74.73 & 89.9 & 77.65 \\ 
        15\% & 57.53 & 83.09 & 72.92 & 89.9 & 75.86 \\ 
        30\% & 55.23 & 76.47 & 71.84 & 89.8 & 73.34 \\ 
        \bottomrule
    \end{tabular}
    \label{tab:noise}
\end{table}
The results show that EigenLoRAx exhibits only minor performance changes even as noise levels increase significantly, indicating some robustness to adapter quality. This suggests that EigenLoRAx can still perform effectively without high quality adapters. However, there is a limit to this robustness. If the signal-to-noise ratio (SNR) in the initial LoRA matrices becomes extremely low—where the LoRAs primarily encode noise rather than meaningful information—the effectiveness of EigenLoRAx diminishes.
In such cases, the principal components (PCs) extracted by EigenLoRAx would correspond to random directions in the parameter space. Consequently, EigenLoRAx’s performance would resemble that of random matrix methods, such as VeRA and NoLA. These methods rely on a large number of random components or bases to approximate meaningful results. While they can achieve reasonable performance, they require fine-tuning a substantially larger number of weights associated with these large number of random components, leading to less efficient learning compared to EigenLoRAx.
This highlights an important consideration: for EigenLoRAx to maintain its efficiency and effectiveness, the initial LoRA matrices must contain at least a minimal level of meaningful signal. This requirement ensures that EigenLoRAx can leverage the structured information encoded in the LoRAs while avoiding the inefficiencies of purely random approaches.

\subsection{Forward pass and backward pass FLOPs}

While it is obvious that EigenLoRAx utilized significantly less number of model parameters as the number of tasks in a domain increase, we show that even in terms of floating point operations on a single task, EigenLoRAx is more efficient than LoRA for our experiments. Even for a single task, the number of floating point operations or multiply-accumulate operations in a forward pass for EigenLoRAx is lower than LoRA for all our experiments. Here are the comparisons of the floating point operations (FLOPs) for the forward (fwd FLOPs) and including backward pass (fwd+bwd FLOPs) for each of the Image Classification and GLUE benchmark (batch size = 1) (MFLOPs - MegaFlops):

\begin{table}[!htb]
    \centering
        \caption{Floating Point Operation calculations for GLUE Benchmark experiment}
    \begin{tabular}{lccc}
    \toprule
        Method & Training Parameters & fwd FLOPs & fwd+bwd FLOPs \\ 
        \midrule
        LoRA & 1.2M & 97,930 MFLOPS & 293,800 MFLOPS \\ 
        VeRA & 25K & 106,390 MFLOPS & 319,170 MFLOPS \\ 
        EigenLoRAx & 12K & 97,030 MFLOPS & 291,080 MFLOPS \\ 
        \bottomrule
    \end{tabular}
    \label{tab:flop_glue}
\end{table}

\begin{table}[!htb]
    \centering
        \caption{Floating Point Operation calculations for Image Classification experiment}
    \begin{tabular}{lccc}
    \toprule
        Method & Training Parameters & fwd FLOPs & fwd+bwd FLOPs \\ 
        \midrule
        LoRA & 36K & 33,773.8 MFLOPS & 101,322 MFLOPS \\ 
        VeRA & 18K & 33,744.8 MFLOPS & 101.234 MFLOPS \\ 
        EigenLoRAx & 96 & 33,730.2 MFLOPS & 101,191 MFLOPS \\ 
        \bottomrule
    \end{tabular}
    \label{tab:flop_class}
\end{table}

\section{Broader Impact and Implications}\label{sec:impact}
This work presents a novel parameter-efficient method for deep learning methods utilizing open source, pretrained Low-Rank Adaptation (LoRA) models. By substantially reducing the computational and memory demands of training and inference, our approach creates a more sustainable and environmentally friendly deep learning paradigm. Our method democratizes accessibility to larger models, making them accessible to researchers and practitioners with limited resources. Furthermore, by harnessing pretrained models, our method can accelerate development and diminish the need for extensive data collection. However, we recognize the inherent risks associated with the use of pretrained models. These include potential biases (racial, gender, etc.), explicit content, since there is no guarantee of the data or method used in training the model, and the potential presence of malicious code. Appropriate caution is advised when using unverified, open-source models.\footnote{\textbf{A Note on Prior Work} This paper reintroduces some ideas initially introduced by the first authors (of this work) in an unpublished work~\citep{kaushik2025eigenlora}. Due to serious issues concerning research ethics, authorship integrity, and academic misconduct, the primary authors of this work (including the first author of both works) chose to independently develop and extend their contributions. These contributions in the previous draft included the original idea and initial algorithm, all experimental results (excluding Section 4.2.2), as well as all ablation studies and additional experiments. The code used for these experiments was also developed solely by the first authors of this work and is publicly available. As the prior work was not accepted or published, self-plagiarism is not a concern. We also acknowledge high-level discussions and exchanges with some of the earlier authors, who, while not directly contributing to the technical developments of this work, provided a limited context that shaped our understanding of the broader problem. We have taken extensive measures to ensure that this paper exclusively reflects the contributions of the authors listed, and any similarities beyond our own contributions are purely coincidental.
\FloatBarrier}

%% file: iclr2025_conference.bib
@article{hu2021lora,
  title={Lora: Low-rank adaptation of large language models},
  author={Hu, Edward J and Shen, Yelong and Wallis, Phillip and Allen-Zhu, Zeyuan and Li, Yuanzhi and Wang, Shean and Wang, Lu and Chen, Weizhu},
  journal={arXiv preprint arXiv:2106.09685},
  year={2021}
}

@inproceedings{
guth2024on,
title={On the universality of neural encodings in {CNN}s},
author={Florentin Guth and Brice M{\'e}nard},
booktitle={ICLR 2024 Workshop on Representational Alignment},
year={2024},
url={https://openreview.net/forum?id=ofEBFOrITI}
}

@misc{
kaushik2025eigenlora,
title={EigenLo{RA}: Recycle trained Adapters for Resource Efficient Adaptation and Inference},
author={Prakhar Kaushik and Aayush Mishra and Ankit Vaidya and Raghavendra Addanki and Ryan A. Rossi and Ani Nenkova and Anqi Liu and Alan Yuille and Jiuxiang Gu},
year={2025},
url={https://openreview.net/forum?id=KxGGZag9gW}
}

@article{zhou2018x,
  title={X-LoRa: An Open Source LPWA Network},
  author={Zhou, Qihao and Zheng, Kan and Hou, Lu and Xing, Jinyu and Xu, Rongtao},
  journal={arXiv preprint arXiv:1812.09012},
  year={2018}
}

@InProceedings{Gain_2020_WACV,
author = {Gain, Alex and Kaushik, Prakhar and Siegelmann, Hava},
title = {Adaptive Neural Connections for Sparsity Learning},
booktitle = {Proceedings of the IEEE/CVF Winter Conference on Applications of Computer Vision (WACV)},
month = {March},
year = {2020}
}

@misc{kaushik2025universalweightsubspacehypothesis,
      title={The Universal Weight Subspace Hypothesis}, 
      author={Prakhar Kaushik and Shravan Chaudhari and Ankit Vaidya and Rama Chellappa and Alan Yuille},
      year={2025},
      eprint={2512.05117},
      archivePrefix={arXiv},
      primaryClass={cs.LG},
      url={https://arxiv.org/abs/2512.05117}, 
}

@InProceedings{Kaushik_2024_CVPR,
    author    = {Kaushik, Prakhar and Kortylewski, Adam and Yuille, Alan},
    title     = {A Bayesian Approach to OOD Robustness in Image Classification},
    booktitle = {Proceedings of the IEEE/CVF Conference on Computer Vision and Pattern Recognition (CVPR)},
    month     = {June},
    year      = {2024},
    pages     = {22988-22997}
}

@misc{chughtai2023toymodeluniversalityreverse,
      title={A Toy Model of Universality: Reverse Engineering How Networks Learn Group Operations}, 
      author={Bilal Chughtai and Lawrence Chan and Neel Nanda},
      year={2023},
      eprint={2302.03025},
      archivePrefix={arXiv},
      primaryClass={cs.LG},
      url={https://arxiv.org/abs/2302.03025}, 
}

@misc{sun2025transformersquaredselfadaptivellms,
      title={Transformer-Squared: Self-adaptive LLMs}, 
      author={Qi Sun and Edoardo Cetin and Yujin Tang},
      year={2025},
      eprint={2501.06252},
      archivePrefix={arXiv},
      primaryClass={cs.LG},
      url={https://arxiv.org/abs/2501.06252}, 
}

@misc{gavish2014optimalhardthresholdsingular,
      title={The Optimal Hard Threshold for Singular Values is 4/sqrt(3)}, 
      author={Matan Gavish and David L. Donoho},
      year={2014},
      eprint={1305.5870},
      archivePrefix={arXiv},
      primaryClass={stat.ME},
      url={https://arxiv.org/abs/1305.5870}, 
}

@article{wu2024mixture,
  title={Mixture of lora experts},
  author={Wu, Xun and Huang, Shaohan and Wei, Furu},
  journal={arXiv preprint arXiv:2404.13628},
  year={2024}
}

@article{Kaushik2024SourceFreeAI,
  title={Source-Free and Image-Only Unsupervised Domain Adaptation for Category Level Object Pose Estimation},
  author={Prakhar Kaushik and Aayush Mishra and Adam Kortylewski and Alan L. Yuille},
  journal={ArXiv},
  year={2024},
  volume={abs/2401.10848},
  url={https://api.semanticscholar.org/CorpusID:267060973}
}

@inproceedings{wang2020NeMo,
title = {NeMo: Neural Mesh Models of Contrastive Features for Robust 3D Pose Estimation},
author = {Angtian, Wang and Kortylewski, Adam and Yuille, Alan},
booktitle = {Proceedings International Conference on Learning Representations (ICLR)},
year = {2021},
}

@article{zhong2024multi,
  title={Multi-lora composition for image generation},
  author={Zhong, Ming and Shen, Yelong and Wang, Shuohang and Lu, Yadong and Jiao, Yizhu and Ouyang, Siru and Yu, Donghan and Han, Jiawei and Chen, Weizhu},
  journal={arXiv preprint arXiv:2402.16843},
  year={2024}
}

@Misc{peft,
  title =        {PEFT: State-of-the-art Parameter-Efficient Fine-Tuning methods},
  author =       {Sourab Mangrulkar and Sylvain Gugger and Lysandre Debut and Younes Belkada and Sayak Paul and Benjamin Bossan},
  howpublished = {\url{https://github.com/huggingface/peft}},
  year =         {2022}
}

@misc{sharma_laser_2023,
	title = {The {Truth} is in {There}: {Improving} {Reasoning} in {Language} {Models} with {Layer}-{Selective} {Rank} {Reduction}},
	shorttitle = {The {Truth} is in {There}},
	url = {http://arxiv.org/abs/2312.13558},
	doi = {10.48550/arXiv.2312.13558},
	urldate = {2024-09-06},
	publisher = {arXiv},
	author = {Sharma, Pratyusha and Ash, Jordan T. and Misra, Dipendra},
	month = dec,
	year = {2023},
	note = {arXiv:2312.13558 [cs]},
}

@misc{meng_pissa_2024,
	title = {{PiSSA}: {Principal} {Singular} {Values} and {Singular} {Vectors} {Adaptation} of {Large} {Language} {Models}},
	shorttitle = {{PiSSA}},
	url = {http://arxiv.org/abs/2404.02948},
	doi = {10.48550/arXiv.2404.02948},
	urldate = {2024-09-08},
	publisher = {arXiv},
	author = {Meng, Fanxu and Wang, Zhaohui and Zhang, Muhan},
	month = may,
	year = {2024},
	note = {arXiv:2404.02948 [cs]},
}

@inproceedings{kopiczko_vera_2023,
	title = {{VeRA}: {Vector}-based {Random} {Matrix} {Adaptation}},
	shorttitle = {{VeRA}},
	url = {https://openreview.net/forum?id=NjNfLdxr3A},
	language = {en},
	urldate = {2024-09-10},
	author = {Kopiczko, Dawid Jan and Blankevoort, Tijmen and Asano, Yuki M.},
	month = oct,
	year = {2023},
}

@book{diao_mixture--domain-adapters_2023,
	title = {Mixture-of-{Domain}-{Adapters}: {Decoupling} and {Injecting} {Domain} {Knowledge} to {Pre}-trained {Language} {Models} {Memories}},
	shorttitle = {Mixture-of-{Domain}-{Adapters}},
	author = {Diao, Shizhe and Xu, Tianyang and Xu, Ruijia and Wang, Jiawei and Zhang, Tong},
	month = jun,
	year = {2023},
	doi = {10.48550/arXiv.2306.05406},
}

@misc{dora,
      title={DoRA: Weight-Decomposed Low-Rank Adaptation}, 
      author={Shih-Yang Liu and Chien-Yi Wang and Hongxu Yin and Pavlo Molchanov and Yu-Chiang Frank Wang and Kwang-Ting Cheng and Min-Hung Chen},
      year={2024},
      eprint={2402.09353},
      archivePrefix={arXiv},
      primaryClass={cs.CL},
      url={https://arxiv.org/abs/2402.09353}, 
}

@misc{kwon2024efficientcompressionoverparameterizeddeep,
      title={Efficient Compression of Overparameterized Deep Models through Low-Dimensional Learning Dynamics}, 
      author={Soo Min Kwon and Zekai Zhang and Dogyoon Song and Laura Balzano and Qing Qu},
      year={2024},
      eprint={2311.05061},
      archivePrefix={arXiv},
      primaryClass={cs.LG},
      url={https://arxiv.org/abs/2311.05061}, 
}

@misc{glue,
      title={GLUE: A Multi-Task Benchmark and Analysis Platform for Natural Language Understanding}, 
      author={Alex Wang and Amanpreet Singh and Julian Michael and Felix Hill and Omer Levy and Samuel R. Bowman},
      year={2019},
      eprint={1804.07461},
      archivePrefix={arXiv},
      primaryClass={cs.CL},
      url={https://arxiv.org/abs/1804.07461}, 
}

@misc{sdxl,
      title={SDXL: Improving Latent Diffusion Models for High-Resolution Image Synthesis}, 
      author={Dustin Podell and Zion English and Kyle Lacey and Andreas Blattmann and Tim Dockhorn and Jonas Müller and Joe Penna and Robin Rombach},
      year={2023},
      eprint={2307.01952},
      archivePrefix={arXiv},
      primaryClass={cs.CV},
      url={https://arxiv.org/abs/2307.01952}, 
}

@article{huang2023lorahub,
  title={Lorahub: Efficient cross-task generalization via dynamic lora composition},
  author={Huang, Chengsong and Liu, Qian and Lin, Bill Yuchen and Pang, Tianyu and Du, Chao and Lin, Min},
  journal={arXiv preprint arXiv:2307.13269},
  year={2023}
}

@misc{jiang2023mistral7b,
      title={Mistral 7B}, 
      author={Albert Q. Jiang and Alexandre Sablayrolles and Arthur Mensch and Chris Bamford and Devendra Singh Chaplot and Diego de las Casas and Florian Bressand and Gianna Lengyel and Guillaume Lample and Lucile Saulnier and Lélio Renard Lavaud and Marie-Anne Lachaux and Pierre Stock and Teven Le Scao and Thibaut Lavril and Thomas Wang and Timothée Lacroix and William El Sayed},
      year={2023},
      eprint={2310.06825},
      archivePrefix={arXiv},
      primaryClass={cs.CL},
      url={https://arxiv.org/abs/2310.06825}, 
}

@misc{brüelgabrielsson2024compressserveservingthousands,
      title={Compress then Serve: Serving Thousands of LoRA Adapters with Little Overhead}, 
      author={Rickard Brüel-Gabrielsson and Jiacheng Zhu and Onkar Bhardwaj and Leshem Choshen and Kristjan Greenewald and Mikhail Yurochkin and Justin Solomon},
      year={2024},
      eprint={2407.00066},
      archivePrefix={arXiv},
      primaryClass={cs.DC},
      url={https://arxiv.org/abs/2407.00066}, 
}

@inproceedings{wang-etal-2022-super,
    title = "Super-{N}atural{I}nstructions: Generalization via Declarative Instructions on 1600+ {NLP} Tasks",
    author = "Wang, Yizhong  and
      Mishra, Swaroop  and
      Alipoormolabashi, Pegah  and
      Kordi, Yeganeh  and
      Mirzaei, Amirreza  and
      Naik, Atharva  and
      Ashok, Arjun  and
      Dhanasekaran, Arut Selvan  and
      Arunkumar, Anjana  and
      Stap, David  and
      Pathak, Eshaan  and
      Karamanolakis, Giannis  and
      Lai, Haizhi  and
      Purohit, Ishan  and
      Mondal, Ishani  and
      Anderson, Jacob  and
      Kuznia, Kirby  and
      Doshi, Krima  and
      Pal, Kuntal Kumar  and
      Patel, Maitreya  and
      Moradshahi, Mehrad  and
      Parmar, Mihir  and
      Purohit, Mirali  and
      Varshney, Neeraj  and
      Kaza, Phani Rohitha  and
      Verma, Pulkit  and
      Puri, Ravsehaj Singh  and
      Karia, Rushang  and
      Doshi, Savan  and
      Sampat, Shailaja Keyur  and
      Mishra, Siddhartha  and
      Reddy A, Sujan  and
      Patro, Sumanta  and
      Dixit, Tanay  and
      Shen, Xudong",
    editor = "Goldberg, Yoav  and
      Kozareva, Zornitsa  and
      Zhang, Yue",
    booktitle = "Proceedings of the 2022 Conference on Empirical Methods in Natural Language Processing",
    month = dec,
    year = "2022",
    address = "Abu Dhabi, United Arab Emirates",
    publisher = "Association for Computational Linguistics",
    url = "https://aclanthology.org/2022.emnlp-main.340/",
    doi = "10.18653/v1/2022.emnlp-main.340",
    pages = "5085--5109",
}

@article{tan2024democratizing,
  title={Democratizing large language models via personalized parameter-efficient fine-tuning},
  author={Tan, Zhaoxuan and Zeng, Qingkai and Tian, Yijun and Liu, Zheyuan and Yin, Bing and Jiang, Meng},
  journal={arXiv preprint arXiv:2402.04401},
  year={2024}
}

@proceedings{acl-2024-long,
    title = "Proceedings of the 62nd Annual Meeting of the Association for Computational Linguistics (Volume 1: Long Papers)",
    editor = "Ku, Lun-Wei  and
      Martins, Andre  and
      Srikumar, Vivek",
    month = aug,
    year = "2024",
    address = "Bangkok, Thailand",
    publisher = "Association for Computational Linguistics",
    url = "https://aclanthology.org/2024.acl-long.0",
}

@inproceedings{
zhou2022mixtureofexperts,
title={Mixture-of-Experts with Expert Choice Routing},
author={Yanqi Zhou and Tao Lei and Hanxiao Liu and Nan Du and Yanping Huang and Vincent Y Zhao and Andrew M. Dai and Zhifeng Chen and Quoc V Le and James Laudon},
booktitle={Advances in Neural Information Processing Systems},
editor={Alice H. Oh and Alekh Agarwal and Danielle Belgrave and Kyunghyun Cho},
year={2022},
url={https://openreview.net/forum?id=jdJo1HIVinI}
}

@misc{zoph2022stmoedesigningstabletransferable,
      title={ST-MoE: Designing Stable and Transferable Sparse Expert Models}, 
      author={Barret Zoph and Irwan Bello and Sameer Kumar and Nan Du and Yanping Huang and Jeff Dean and Noam Shazeer and William Fedus},
      year={2022},
      eprint={2202.08906},
      archivePrefix={arXiv},
      primaryClass={cs.CL},
      url={https://arxiv.org/abs/2202.08906}, 
}

@misc{zhang2023adaloraadaptivebudgetallocation,
      title={AdaLoRA: Adaptive Budget Allocation for Parameter-Efficient Fine-Tuning}, 
      author={Qingru Zhang and Minshuo Chen and Alexander Bukharin and Nikos Karampatziakis and Pengcheng He and Yu Cheng and Weizhu Chen and Tuo Zhao},
      year={2023},
      eprint={2303.10512},
      archivePrefix={arXiv},
      primaryClass={cs.CL},
      url={https://arxiv.org/abs/2303.10512}, 
}

@inproceedings{
nola,
title={{NOLA}: Compressing Lo{RA} using Linear Combination of Random Basis},
author={Soroush Abbasi Koohpayegani and Navaneet K L and Parsa Nooralinejad and Soheil Kolouri and Hamed Pirsiavash},
booktitle={The Twelfth International Conference on Learning Representations},
year={2024},
url={https://openreview.net/forum?id=TjfXcDgvzk}
}

@misc{kaushik2021understandingcatastrophicforgettingremembering,
      title={Understanding Catastrophic Forgetting and Remembering in Continual Learning with Optimal Relevance Mapping}, 
      author={Prakhar Kaushik and Alex Gain and Adam Kortylewski and Alan Yuille},
      year={2021},
      eprint={2102.11343},
      archivePrefix={arXiv},
      primaryClass={cs.LG},
      url={https://arxiv.org/abs/2102.11343}, 
}

@inproceedings{
vision_transformer,
title={An Image is Worth 16x16 Words: Transformers for Image Recognition at Scale},
author={Alexey Dosovitskiy and Lucas Beyer and Alexander Kolesnikov and Dirk Weissenborn and Xiaohua Zhai and Thomas Unterthiner and Mostafa Dehghani and Matthias Minderer and Georg Heigold and Sylvain Gelly and Jakob Uszkoreit and Neil Houlsby},
booktitle={International Conference on Learning Representations},
year={2021},
url={https://openreview.net/forum?id=YicbFdNTTy}
}

@article{diffusion,
  title={High-Resolution Image Synthesis with Latent Diffusion Models},
  author={Robin Rombach and A. Blattmann and Dominik Lorenz and Patrick Esser and Bj{\"o}rn Ommer},
  journal={2022 IEEE/CVF Conference on Computer Vision and Pattern Recognition (CVPR)},
  year={2021},
  pages={10674-10685},
  url={https://api.semanticscholar.org/CorpusID:245335280}
}

@article{luo2023towards,
  title={Towards Efficient Visual Adaption via Structural Re-parameterization},
  author={Luo, Gen and Huang, Minglang and Zhou, Yiyi  and Sun, Xiaoshuai and Jiang, Guangnan and Wang, Zhiyu and Ji, Rongrong},
  journal={arXiv preprint arXiv:2302.08106},
  year={2023}
}

@article{Chen2022AdaptFormerAV,
  title={AdaptFormer: Adapting Vision Transformers for Scalable Visual Recognition},
  author={Shoufa Chen and Chongjian Ge and Zhan Tong and Jiangliu Wang and Yibing Song and Jue Wang and Ping Luo},
  journal={ArXiv},
  year={2022},
  volume={abs/2205.13535},
  url={https://api.semanticscholar.org/CorpusID:249097890}
}

@InProceedings{pmlr-v97-houlsby19a,
  title = 	 {Parameter-Efficient Transfer Learning for {NLP}},
  author =       {Houlsby, Neil and Giurgiu, Andrei and Jastrzebski, Stanislaw and Morrone, Bruna and De Laroussilhe, Quentin and Gesmundo, Andrea and Attariyan, Mona and Gelly, Sylvain},
  booktitle = 	 {Proceedings of the 36th International Conference on Machine Learning},
  pages = 	 {2790--2799},
  year = 	 {2019},
  editor = 	 {Chaudhuri, Kamalika and Salakhutdinov, Ruslan},
  volume = 	 {97},
  series = 	 {Proceedings of Machine Learning Research},
  month = 	 {09--15 Jun},
  publisher =    {PMLR},
  url = 	 {https://proceedings.mlr.press/v97/houlsby19a.html}
}

@inproceedings{tune1,
    title = "The Power of Scale for Parameter-Efficient Prompt Tuning",
    author = "Lester, Brian  and
      Al-Rfou, Rami  and
      Constant, Noah",
    editor = "Moens, Marie-Francine  and
      Huang, Xuanjing  and
      Specia, Lucia  and
      Yih, Scott Wen-tau",
    booktitle = "Proceedings of the 2021 Conference on Empirical Methods in Natural Language Processing",
    month = nov,
    year = "2021",
    address = "Online and Punta Cana, Dominican Republic",
    publisher = "Association for Computational Linguistics",
    url = "https://aclanthology.org/2021.emnlp-main.243",
    doi = "10.18653/v1/2021.emnlp-main.243",
    pages = "3045--3059",
}

@inproceedings{tune2,
    title = "Residual Prompt Tuning: improving prompt tuning with residual reparameterization",
    author = "Razdaibiedina, Anastasiia  and
      Mao, Yuning  and
      Khabsa, Madian  and
      Lewis, Mike  and
      Hou, Rui  and
      Ba, Jimmy  and
      Almahairi, Amjad",
    editor = "Rogers, Anna  and
      Boyd-Graber, Jordan  and
      Okazaki, Naoaki",
    booktitle = "Findings of the Association for Computational Linguistics: ACL 2023",
    month = jul,
    year = "2023",
    address = "Toronto, Canada",
    publisher = "Association for Computational Linguistics",
    url = "https://aclanthology.org/2023.findings-acl.421",
    doi = "10.18653/v1/2023.findings-acl.421",
    pages = "6740--6757",
}

@article{tune3,
  title={Prompt tuning for parameter-efficient medical image segmentation},
  author={Fischer, Marc and Bartler, Alexander and Yang, Bin},
  journal={Medical Image Analysis},
  volume={91},
  pages={103024},
  year={2024},
  publisher={Elsevier}
}

@article{Ahmed2020TheDO,
  title={The De-democratization of AI: Deep Learning and the Compute Divide in Artificial Intelligence Research},
  author={Nuri Mahmoud Ahmed and Muntasir Wahed},
  journal={ArXiv},
  year={2020},
  volume={abs/2010.15581},
  url={https://api.semanticscholar.org/CorpusID:225102971}
}

@article{Besiroglu2024TheCD,
  title={The Compute Divide in Machine Learning: A Threat to Academic Contribution and Scrutiny?},
  author={Tamay Besiroglu and Sage Andrus Bergerson and Amelia Michael and Lennart Heim and Xueyun Luo and Neil Thompson},
  journal={ArXiv},
  year={2024},
  volume={abs/2401.02452},
  url={https://api.semanticscholar.org/CorpusID:266818226}
}

@article{edge,
author = {Liu, Di and Kong, Hao and Luo, Xiangzhong and Liu, Weichen and Subramaniam, Ravi},
title = {Bringing AI to edge: From deep learning’s perspective},
year = {2022},
issue_date = {May 2022},
publisher = {Elsevier Science Publishers B. V.},
address = {NLD},
volume = {485},
number = {C},
issn = {0925-2312},
url = {https://doi.org/10.1016/j.neucom.2021.04.141},
doi = {10.1016/j.neucom.2021.04.141},
journal = {Neurocomput.},
month = may,
pages = {297–320},
numpages = {24}
}

@article{Wu2021SustainableAE,
  title={Sustainable AI: Environmental Implications, Challenges and Opportunities},
  author={Carole-Jean Wu and Ramya Raghavendra and Udit Gupta and Bilge Acun and Newsha Ardalani and Kiwan Maeng and Gloria Chang and Fiona Aga Behram and James Huang and Charles Bai and Michael K. Gschwind and Anurag Gupta and Myle Ott and Anastasia Melnikov and Salvatore Candido and David Brooks and Geeta Chauhan and Benjamin Lee and Hsien-Hsin S. Lee and Bugra Akyildiz and Maximilian Balandat and Joe Spisak and Ravi Kumar Jain and Michael G. Rabbat and Kim M. Hazelwood},
  journal={ArXiv},
  year={2021},
  volume={abs/2111.00364},
  url={https://api.semanticscholar.org/CorpusID:240354766}
}

@article{Ligozat2021UnravelingTH,
  title={Unraveling the hidden environmental impacts of AI solutions for environment},
  author={Anne-Laure Ligozat and Julien Lef{\`e}vre and Aur{\'e}lie Bugeau and Jacques Combaz},
  journal={ArXiv},
  year={2021},
  volume={abs/2110.11822},
  url={https://api.semanticscholar.org/CorpusID:239616423}
}

@article{Burer2003ANP,
  title={A nonlinear programming algorithm for solving semidefinite programs via low-rank factorization},
  author={Samuel Burer and Renato D. C. Monteiro},
  journal={Mathematical Programming},
  year={2003},
  volume={95},
  pages={329-357},
  url={https://api.semanticscholar.org/CorpusID:7691228}
}

@Inbook{Ma2024,
author="Ma, Cong
and Xu, Xingyu
and Tong, Tian
and Chi, Yuejie",
title="Provably Accelerating Ill-Conditioned Low-Rank Estimation via Scaled Gradient Descent, Even with Overparameterization",
bookTitle="Explorations in the Mathematics of Data Science: The Inaugural Volume of the Center for Approximation and Mathematical Data Analytics",
year="2024",
publisher="Springer Nature Switzerland",
address="Cham",
pages="133--165",
isbn="978-3-031-66497-7",
doi="10.1007/978-3-031-66497-7_7",
url="https://doi.org/10.1007/978-3-031-66497-7_7"
}

@ARTICLE{chi19_low,
  author={Chi, Yuejie and Lu, Yue M. and Chen, Yuxin},
  journal={IEEE Transactions on Signal Processing}, 
  title={Nonconvex Optimization Meets Low-Rank Matrix Factorization: An Overview}, 
  year={2019},
  volume={67},
  number={20},
  pages={5239-5269},
  doi={10.1109/TSP.2019.2937282}}

@inproceedings{browngpt3,
author = {Brown, Tom B. and Mann, Benjamin and Ryder, Nick and Subbiah, Melanie and Kaplan, Jared and Dhariwal, Prafulla and Neelakantan, Arvind and Shyam, Pranav and Sastry, Girish and Askell, Amanda and Agarwal, Sandhini and Herbert-Voss, Ariel and Krueger, Gretchen and Henighan, Tom and Child, Rewon and Ramesh, Aditya and Ziegler, Daniel M. and Wu, Jeffrey and Winter, Clemens and Hesse, Christopher and Chen, Mark and Sigler, Eric and Litwin, Mateusz and Gray, Scott and Chess, Benjamin and Clark, Jack and Berner, Christopher and McCandlish, Sam and Radford, Alec and Sutskever, Ilya and Amodei, Dario},
title = {Language models are few-shot learners},
year = {2020},
isbn = {9781713829546},
publisher = {Curran Associates Inc.},
address = {Red Hook, NY, USA},
booktitle = {Proceedings of the 34th International Conference on Neural Information Processing Systems},
articleno = {159},
numpages = {25},
location = {Vancouver, BC, Canada},
series = {NIPS '20}
}

@article{roberta,
  title={RoBERTa: A Robustly Optimized BERT Pretraining Approach},
  author={Yinhan Liu and Myle Ott and Naman Goyal and Jingfei Du and Mandar Joshi and Danqi Chen and Omer Levy and Mike Lewis and Luke Zettlemoyer and Veselin Stoyanov},
  journal={ArXiv},
  year={2019},
  volume={abs/1907.11692},
  url={https://api.semanticscholar.org/CorpusID:198953378}
}

@inproceedings{food101,
  title = {Food-101 -- Mining Discriminative Components with Random Forests},
  author = {Bossard, Lukas and Guillaumin, Matthieu and Van Gool, Luc},
  booktitle = {European Conference on Computer Vision},
  year = {2014}
}

@InProceedings{flowers102,
  author       = "Maria-Elena Nilsback and Andrew Zisserman",
  title        = "Automated Flower Classification over a Large Number of Classes",
  booktitle    = "Indian Conference on Computer Vision, Graphics and Image Processing",
  month        = "Dec",
  year         = "2008",
}

@article{cifar100,
title= {CIFAR-100 (Canadian Institute for Advanced Research)},
journal= {},
author= {Alex Krizhevsky and Vinod Nair and Geoffrey Hinton},
year= {2009},
url= {http://www.cs.toronto.edu/~kriz/cifar.html},
terms= {}
}

@misc{von-platen-etal-2022-diffusers,
  author = {Patrick von Platen and Suraj Patil and Anton Lozhkov and Pedro Cuenca and Nathan Lambert and Kashif Rasul and Mishig Davaadorj and Dhruv Nair and Sayak Paul and William Berman and Yiyi Xu and Steven Liu and Thomas Wolf},
  title = {Diffusers: State-of-the-art diffusion models},
  year = {2022},
  publisher = {GitHub},
  journal = {GitHub repository},
  howpublished = {\url{https://github.com/huggingface/diffusers}}
}

@article{federated,
  title={Asynchronous Federated Continual Learning},
  author={Donald Shenaj and Marco Toldo and Alberto Rigon and Pietro Zanuttigh},
  journal={2023 IEEE/CVF Conference on Computer Vision and Pattern Recognition Workshops (CVPRW)},
  year={2023},
  pages={5055-5063},
  url={https://api.semanticscholar.org/CorpusID:258041245}
}

@article{bartlett2003rademacher,
author = {Bartlett, Peter L. and Mendelson, Shahar},
title = {Rademacher and gaussian complexities: risk bounds and structural results},
year = {2003},
issue_date = {3/1/2003},
publisher = {JMLR.org},
volume = {3},
number = {null},
issn = {1532-4435},
journal = {J. Mach. Learn. Res.},
month = mar,
pages = {463–482},
numpages = {20}
}
